%% file: main.tex
\documentclass[11pt, a4paper, logo]{googlecloud}

\pdftrailerid{redacted}

\makeatletter
\renewcommand\bibentry[1]{\nocite{#1}{\frenchspacing\@nameuse{BR@r@#1\@extra@b@citeb}}}
\makeatother

\RequirePackage{algorithm}
\RequirePackage{algorithmic}

\usepackage[authoryear, sort&compress, round]{natbib}

\usepackage{tcolorbox}
\usepackage{times}
\usepackage{latexsym}
\usepackage{kantlipsum, lipsum}
\usepackage{dsfont}
\usepackage{makecell}
\usepackage{url}            % simple URL typesetting
\usepackage{nicefrac}       % compact symbols for 1/2, etc.
\usepackage{multicol}
\usepackage{bbm}
\usepackage{multirow}
\usepackage{adjustbox}
\usepackage{listings}
\usepackage{soul}
\usepackage{float}
\usepackage{wrapfig}
\usepackage{blindtext}
\usepackage{tablefootnote}
\usepackage{amsfonts}
\usepackage[flushleft]{threeparttable}
\usepackage{bbding}
\usepackage{xcolor}
\usepackage{xspace}
\usepackage{bm}
\usepackage{enumitem}
\usepackage{setspace}
\usepackage{color}
\usepackage{longtable}
\usepackage[normalem]{ulem}
\usepackage{ulem}
\usepackage[nomargin,inline,marginclue,draft]{fixme}
\usepackage{balance}
\usepackage{verbatim}
\usepackage{diagbox}
\usepackage{changepage}
\usepackage{pifont}
\usepackage{array}   % Optional: Improves table column alignment
\input{math_commands.tex}

\usepackage{microtype}
\usepackage{graphicx}
\usepackage{subcaption}
\usepackage{booktabs} % for professional tables
\usepackage{arydshln} % must load after booktabs to avoid \midrule conflict in longtable

\usepackage{hyperref}

\usepackage{amsmath}
\usepackage{amssymb}
\usepackage{mathtools}
\usepackage{amsthm}

\usepackage[capitalize,noabbrev]{cleveref}

\theoremstyle{plain}

\theoremstyle{definition}

\theoremstyle{remark}

\usepackage[textsize=tiny]{todonotes}

\newcommand{\coe}{Chain-of-Evidence}
\newcommand{\coeaudit}{CoE Integrity Audit}

\newcommand{\sys}{\texttt{\textbf{ScientistOne}}}

\usepackage{xcolor}

\usepackage{pifont}
\usepackage[table]{xcolor}

\usepackage{minted} 
\setminted{tabsize=2, breaklines=true, breakanywhere=true}
\tcbuselibrary{skins,breakable}
\newtcolorbox{promptbox}[1]{
  colback=white,
  colframe=blue!100,
  colbacktitle=blue!10,
  coltitle=black,
  title=#1,
  fonttitle=\bfseries\sffamily,
  breakable,
  enhanced,
  boxrule=0.8pt
}
\usepackage{listings}
\title{\sys{}: Towards Human-Level Autonomous Research via Chain-of-Evidence}

\correspondingauthor{}

\makeatletter
\renewcommand\AB@authnote[1]{}
\renewcommand\AB@affilnote[1]{}
\makeatother

\author{Rui Meng}
\author{Bhavana Dalvi Mishra*}
\author{Jiefeng Chen*}
\author{Chun-Liang Li}
\author{Palash Goyal}
\author{Mihir Parmar}
\author{Yiwen Song}
\author{Yale Song}
\author{Rajarishi Sinha}
\author{Parthasarathy Ranganathan}
\author{Burak Gokturk}
\author{Jinsung Yoon}
\author{Tomas Pfister}

\affil{Google Cloud AI Research}

\input{sections/00_abstract}

\begin{document}

\maketitle
\def\thefootnote{*}\footnotetext{These authors contributed equally to this work.}\def\thefootnote{\arabic{footnote}}

\input{sections/00_abstract}
\input{sections/01_introduction}
\input{sections/02_related_work}
\input{sections/03_coe_standard}
\input{sections/04_system}

\input{sections/05_coe_audit}
\input{sections/06_experiments}

\input{sections/07_generalizability}
\input{sections/010_conclusion}
\input{sections/011_limitations}

% \section*{Acknowledgements}

\bibliographystyle{abbrvnat}
\nobibliography*
\bibliography{references}

%%%%%%%%%%%%%%%%%%%%%%%%%%%%%%%%%%%%%%%%%%%%%%%%%%%%%%%%%%%%%%%%%%%%%%%%%%%%%%%
% APPENDIX
%%%%%%%%%%%%%%%%%%%%%%%%%%%%%%%%%%%%%%%%%%%%%%%%%%%%%%%%%%%%%%%%%%%%%%%%%%%%%%%
\newpage
\appendix
\input{sections/012_appendix}

\end{document}

%% file: math_commands.tex
\usepackage{amsmath,amsfonts,bm}

\def\eqref#1{equation~\ref{#1}}
\def\1{\bm{1}}

\DeclareMathAlphabet{\mathsfit}{\encodingdefault}{\sfdefault}{m}{sl}
\SetMathAlphabet{\mathsfit}{bold}{\encodingdefault}{\sfdefault}{bx}{n}

%% file: sections/00_abstract.tex
\begin{abstract}
Autonomous research agents produce competitive solutions and professional-looking manuscripts, yet their outputs can contain verifiability failures undetectable by evaluations that only assess surface presentation rather than evidence grounding: fabricated citations, unreproducible scores, and method descriptions that diverge from the implementation.
These failures share a common root: no existing evaluation protocol audits whether claims are supported, and no existing autonomous research system is designed to trace claims back to evidence.
We address this gap through three contributions. First, \emph{\coe{}} (CoE), a verifiability framework requiring every claim to be traceable to its evidence source. Second, \sys{}, an end-to-end autonomous research system that maintains evidence chains by construction throughout literature review, solution discovery, and paper writing. Third, \emph{\coeaudit{}}, a post-hoc audit whose four integrity checks---score verification, specification violation, reference verification, and method--code alignment---apply uniformly to all systems.
Across 75~papers spanning five systems and five frontier research tasks, we find that every baseline exhibits at least one systematic failure mode: hallucinated reference rates reach 21\%, score verification passes in as few as 42\% of papers, and method--code alignment ranges from 20\% to 80\%.
\sys{} is the only system to achieve zero hallucinated references (0/337 bibliography entries), perfect score verification (12/12), and the highest method--code alignment (14/15), while matching or exceeding human expert performance on all five tasks.
We further demonstrate that \sys{} generalizes to six additional tasks spanning medical imaging, fine-grained recognition, 3D perception, and parameter-constrained language modeling, achieving state-of-the-art on Parameter Golf and gold medals on MLE-Bench tasks where baselines fail entirely.
Project website: \url{https://scientist-one.github.io/}
\end{abstract}

%% file: sections/01_introduction.tex
%% ============================================================
\section{Introduction}
\label{sec:intro}
%% ============================================================

%% --- Domain landscape ---

Large language models are increasingly deployed not as isolated assistants but as autonomous agents that conduct entire research workflows---from literature review and hypothesis generation through experimental design and execution to manuscript writing~\citep{lu2024aiscientist, yamada2025aiscientistv2, huang2025deepscientist, tang2025airesearcher, schmidgall2025agentlab, jansen2025codescientist}.
On systems-optimization tasks, such agents now produce solutions competitive with human experts~\citep{cheng2025barbarians, novikov2025alphaevolve}, and end-to-end pipelines have generated papers accepted at peer-reviewed workshops~\citep{yamada2025aiscientistv2}.
The resulting artifacts---code, experimental results, and professional-looking manuscripts---are increasingly difficult to distinguish from human-authored research on surface quality alone.

%% --- Problem buildup: the verifiability gap ---

This rapid capability growth exposes a structural tension between \emph{generation} and \emph{verification}.
Autonomous research systems operate as multi-stage pipelines in which each stage consumes the output of the previous one: a literature summary shapes the hypothesis, the hypothesis determines the experiment, and experimental results feed into the manuscript.
In such architectures, errors introduced at any stage are not merely preserved but amplified---a flawed summary can bias experimental design, and a misinterpreted result can carry through into a paper that appears internally coherent, precisely because the same error is reflected consistently across sections.
The risk grows with trajectory length: agents struggle to track an ever-expanding context \citep{liu2024lost,liu2023agentbench}, hallucinate, and drift from the original objective.
The problem is exacerbated by fundamental limitations in how language models handle evidence: generated text is difficult to verify against sources~\citep{liu2023evaluating}, factual claims drift from their grounding~\citep{min2023factscore}, and scientific citations are frequently inaccurate or fabricated~\citep{press2024citeme}.

In autonomous pipelines, these failure modes interact and compound---a model can overstate method descriptions beyond what the code implements, report scores that do not reproduce under the benchmark's own evaluator, and populate bibliographies from parametric memory rather than retrieval, all while producing text that reads as technically sound.
Existing evaluation protocols, whether automated review scores or benchmark leaderboards, assess surface presentation (i.e., how the paper reads) and procedural completion but do not check whether individual claims trace to supporting evidence.

%% --- Condensed motivation from our audit ---

This verifiability gap is not hypothetical.
In a systematic audit of 75~papers from five autonomous research systems across five benchmark tasks, we find that \emph{every baseline system exhibits evidence chain failures}: hallucinated references that do not correspond to any real publication (up to 21\% of all bibliography entries), method sections that describe algorithms not present in the submitted code, unreproducible scores, and solution code that exploits the evaluator rather than solving the task.
These failures share a common root cause: \emph{no existing evaluation protocol audits whether claims are supported, and no existing autonomous research system is designed to trace claims back to evidence.}

%% --- Our answer: CoE ---

We address this with \emph{\coe{}} (CoE), a verifiability framework for AI-driven research.
Just as ACID\footnote{Atomicity, consistency, isolation, durability.}~\citep{haerder1983principles} defines what ``reliable'' means for a database transaction,
CoE defines what ``verifiable'' means for a research claim: \textbf{every claim must trace, through a recorded evidence chain, to a grounding source.}
We instantiate CoE in three ways:

\begin{enumerate}[nosep,leftmargin=*]
  \item \textbf{The CoE Standard} (\S\ref{sec:coe}): a claim taxonomy (citation, numerical, methodological, conclusion) and the evidence chain structure required for each type.

  \item \textbf{\sys{}} (\S\ref{sec:system}): an end-to-end autonomous research system whose pipeline---Problem Investigator, Discovery Engine, and Paper Writer with Claim Verifier---is designed to satisfy CoE natively.
  The Problem Investigator reads up to 100 full-text PDFs per topic, producing grounded experiment briefs. And the Claim Verifier checks every claim in the draft against its declared evidence source before the final paper is produced.

  \item \textbf{\coeaudit{}} (\S\ref{sec:coeaudit}): a post-hoc audit for verifying an AI-driven research paper through four integrity checks---Score Verification, Specification Violation, Reference Verification, and Method-Code Alignment---targeting the most damaging evidence chain failures.
\end{enumerate}

We apply \coeaudit{} to 15~papers from each of five systems across five frontier systems-research tasks from ADRS~\citep{cheng2025barbarians,skydiscover2026}~(\S\ref{sec:experiments}).
Every baseline exhibits at least one integrity check failure.
\sys{} achieves zero hallucinated references (0/337 bibliography entries), perfect score verification (12/12), and the highest method--code alignment (14/15), while matching or exceeding human expert solver performance on all five tasks.
We further demonstrate that \sys{} generalizes to six additional tasks spanning medical imaging, fine-grained recognition, 3D perception, and parameter-constrained language modeling, achieving state-of-the-art on Parameter Golf and gold medals on MLE-Bench tasks where baselines fail entirely.

%% file: sections/02_related_work.tex
%% ============================================================
\section{Related Work}
\label{sec:related}
%% ============================================================

\paragraph{Autonomous research agents.}
End-to-end autonomous research systems have rapidly expanded from constrained ML templates to multi-stage pipelines that coordinate literature grounding, hypothesis generation, experimentation, and paper writing.
The AI Scientist~\citep{lu2024aiscientist} pioneered end-to-end automation but operates on fixed ML templates with frequent hallucinations in writing and limited paper quality.
AI Scientist-v2~\citep{yamada2025aiscientistv2} advances this with best-first tree search (BFTS) over experimental branches and review-aware reporting, achieving workshop-level paper quality.
Concurrent systems extend the pipeline in different directions.
On the ideation side, PiFlow~\citep{wu2025piflow} steers hypothesis exploration via information-theoretic principle selection and CodeScientist~\citep{jansen2025codescientist} grounds ideation jointly in literature and code.
Curie~\citep{kon2025curie} validates experimental execution through reproducibility checks analogous to our I1 Score Verification, though it does not audit whether written claims faithfully reflect the validated results.
Agent Laboratory~\citep{schmidgall2025agentlab} introduces human gating into the pipeline. AlphaEvolve~\citep{novikov2025alphaevolve} applies evolutionary search to algorithmic optimization, and EvoScientist~\citep{li2026evoscientist} uses multi-agent self-evolution for end-to-end discovery.
We evaluate AI Scientist-v2 alongside three additional systems---AutoResearchClaw~\citep{li2024autoresearchclaw}, DeepScientist~\citep{huang2025deepscientist}, and AI-Researcher~\citep{tang2025airesearcher}---whose architectural choices produce distinct integrity profiles (\S\ref{sec:exp:integrity}).
Despite this architectural diversity, a common pattern emerges: generation and execution capabilities have scaled faster than validation and provenance mechanisms, so systems that produce professional-looking manuscripts may still contain broken evidence chains.
\sys{} targets this gap---rather than advancing the autonomy frontier, we focus on making autonomous research outputs verifiable.

\paragraph{LLM-driven optimization and benchmarks.}
The ADRS benchmark~\citep{cheng2025barbarians} collects real frontier computer system research questions and serves as our primary evaluation testbed.
EvoX~\citep{liu2026evox} and AdaEvolve~\citep{cemri2026adaevolve} achieve strong results on ADRS by focusing on algorithm discovery and implementation optimization without literature grounding or paper writing.
Broader evaluation resources have recently proliferated. Auto-Bench~\citep{zhong2025autobench}, ResearchBench~\citep{liu2025researchbench}, and ResearcherBench~\citep{xu2025researcherbench} evaluate research-adjacent capabilities such as causal reasoning, hypothesis generation, and research question answering. MLAgentBench~\citep{huang2024mlagentbench}, EXP-Bench~\citep{weng2025expbench}, and PaperBench~\citep{starace2025paperbench} stress-test experimentation, replication, and execution reliability. AIRS-Bench~\citep{jansen2025airsbench} tests agent performance on tasks drawn from published ML papers. FIRE-Bench~\citep{shojaee2025firebench} evaluates whether agents can rediscover established findings through full-cycle experimentation.
However, most benchmarks measure discovery performance---whether a system can produce competitive solutions---rather than whether the resulting claims are actually supported by evidence.

\paragraph{Scientific integrity and provenance.}
Current autonomous research systems produce written outputs with varying degrees of traceability: direct manuscript drafting where an LLM generates prose from agent outputs~\citep{lu2024aiscientist,jansen2025codescientist,tang2025airesearcher}, and review-aware revision where reviewer feedback refines the manuscript~\citep{yamada2025aiscientistv2}.
Both approaches produce fluent papers but lack mechanisms to ensure that reported numbers trace to specific execution artifacts, masking broken evidence chains.
Prior work on citation verifiability~\citep{liu2023evaluating}, factual accuracy~\citep{min2023factscore}, and citation attribution~\citep{press2024citeme} performs post-hoc detection at the text level.
CoE differs in two ways: it defines verifiability at the level of individual claims (each must trace to a grounding source through the full research artifact), and it covers paper, code, and evaluator logs jointly, not just text.
\coeaudit{} operationalizes this standard as a cross-system audit, subject to the artifact requirements detailed in \S\ref{sec:coeaudit}.

%% file: sections/03_coe_standard.tex
%% ============================================================
\section{Chain-of-Evidence: A Standard for Research Verifiability}
\label{sec:coe}
%% ============================================================

\begin{quote}
\emph{\textbf{Principle:} Every claim produced by a research system must be traceable, through a recorded chain of supporting claims and evidence, to a grounding source.}
\end{quote}

\noindent A credible research claim must be backed by verifiable evidence.
Without this requirement, the same system that produces a plausible-sounding paper can also produce fabricated citations, hallucinated numbers, and descriptions of experiments that never happened.
Just as a database that violates ACID may return plausible-looking query results even as it silently corrupts data---a transfer debits one account but never credits another, yet both balances look valid---a research system that violates CoE may produce plausible-looking papers whose claims cannot be traced to evidence---the paper reads well, but the scores do not reproduce.
ACID does not prescribe \emph{how} to build a database. It prescribes what properties the database must have.
CoE plays the same role for research artifacts.

We define four primary claim types, each with a required evidence chain shape. The taxonomy is not exhaustive but covers the claim types that are tractably verifiable with current tools---other types (e.g., qualitative observations, theoretical properties) require domain expertise or subjective judgment that is harder to automate.
\textbf{\emph{Citation claims}} (e.g., ``Smith et al.\ showed X'') require that the cited work exists in a scholarly database and that its content is consistent with how it is described in the paper.
\textbf{\emph{Numerical claims}} (e.g., ``achieves 87.3\% on Prism'') must trace from the reported value to a recorded output (e.g., an execution log, experimental measurement, or simulation result).
\textbf{\emph{Methodological claims}} (e.g., ``we use a 3-layer MLP'') must resolve from the method description to the corresponding implementation.
\textbf{\emph{Conclusion claims}} (e.g., ``outperforms baseline by 5\%'') must derive from supporting claims---numerical, methodological, or both---through verifiable reasoning.
CoE is deliberately architecture-agnostic: it defines what properties a verifiable artifact should have, not how the system should construct one.
The standard is also author-agnostic---the same evidence chains are required whether a paper is human- or machine-authored---but we focus on autonomous systems because their failure modes are systematic and rapidly growing in scale.

In the following sections, we describe \sys{}, an autonomous research system designed to satisfy CoE by construction (\S\ref{sec:system}), and \coeaudit{}, a post-hoc audit that measures how well any system's artifacts meet the standard through four integrity checks (\S\ref{sec:coeaudit}).

%% file: sections/04_system.tex
%% ============================================================
\section{\sys{}: Research with Verifiability}
\label{sec:system}
%% ============================================================

\begin{figure}[t]
\centering
\includegraphics[width=\textwidth]{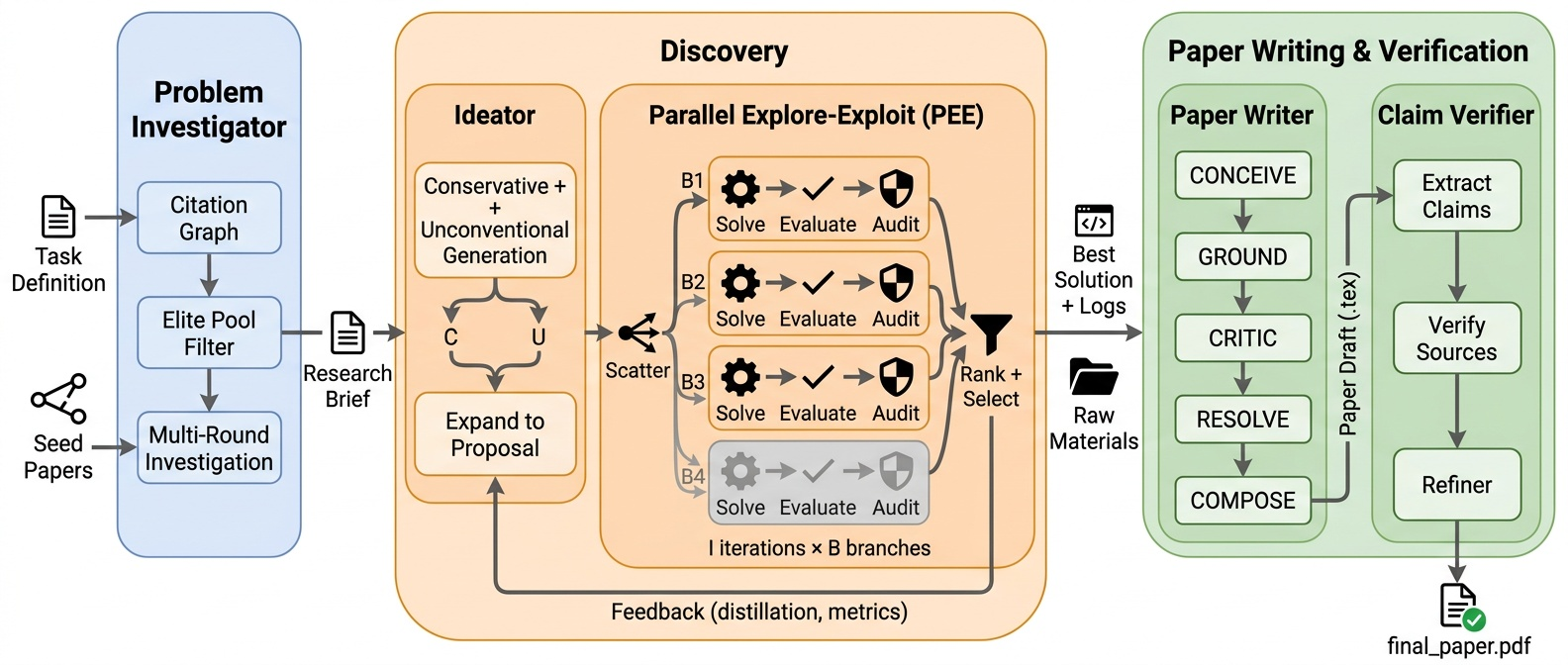}
\caption{\sys{} pipeline. Stage~1 grounds the literature via retrieved PDFs; Stage~2 explores and evaluates solutions across parallel branches; Stage~3 writes and verifies the paper, with a Claim Verifier that checks every claim against its evidence source before the final output is produced.}
\label{fig:architecture}
\end{figure}

We now describe \sys{}, an end-to-end autonomous research system whose three-stage architecture is shaped by the CoE requirements: each module is designed to produce structured artifacts that carry the provenance metadata needed to verify claims against their evidence (Figure~\ref{fig:architecture}).

\subsection{Stage 1: Literature Grounding}
\label{sec:system:pi}

The Problem Investigator (PI) is designed to ensure that every paper the system cites was retrieved from a scholarly database, read in full text, and recorded with provenance metadata.
Without structured retrieval, autonomous systems tend to generate citations from model memory---in our audit, systems without retrieval-grounded references exhibit hallucinated reference rates of up to 21\% (\S\ref{sec:exp:integrity}).
PI addresses this by construction: starting from seed papers, it builds a citation graph via scholarly database queries, reads up to 100 full-text PDFs per topic, and produces a structured research brief.
The brief feeds the Ideator, and PI's seed reference bibliography provides grounding material for citation claims in the final paper.
Pipeline details are in Appendix~\ref{app:system_details}.

\subsection{Stage 2: Discovery}
\label{sec:system:discovery}

The Ideator generates candidate approaches based on the PI brief, scores them on novelty and feasibility, and distributes the top-ranked proposals across parallel branches of the \emph{Parallel Explore-Exploit} (PEE) orchestrator.
Each branch runs an isolated cycle: a Solver agent iterates up to $E$ evaluated versions per node, with a task-specific evaluator scoring each submission.
At each iteration, the top-$K$ branches are retained, and the remaining slots are filled with new branches derived from these top performers via fresh ideation.
After $I$ iterations across $B$ branches, a best-run selector filters out solutions flagged for specification violations (Section~\ref{appx:i2_spec_violation}), selects the highest-scoring remaining solution, and runs ablation experiments on it.
The evaluator scores, execution logs, and ablation results are passed to Stage~3 as source material for paper writing and claim verification.
Architecture details are in Appendix~\ref{app:system_details}.

\subsection{Stage 3: Paper Writing \& Verification}
\label{sec:system:stage3}

\input{sections/04.5_system_claim_writer}

%% file: sections/04.5_system_claim_writer.tex
\paragraph{Paper Writer.}
\label{sec:paper_writer}
\label{sec:system:writer}

The Paper Writer produces \LaTeX{} through a five-stage claim-grounded pipeline.
\textsc{Conceive} reads all assembled raw materials---PI brief, experimental log, verified scores, solver code, and seed-paper abstracts---and emits a \emph{research representation}: a markdown narrative where every factual claim carries an inline evidence tag binding it to a specific workspace artifact (a log line number, a score file entry, a citation key, or an ablation result).
\textsc{Ground} then validates each tag deterministically: the reported score must match the best-run score from discovery, baselines must be traceable to PI brief entries or marked \textsc{estimated}, and every referenced artifact must exist.
\textsc{Critic} audits what deterministic checks cannot---gap--approach alignment, internal contradictions, overclaims, missing comparisons, and baseline fairness---returning \textsc{pass} or a list of issues.
\textsc{Resolve} rewrites the representation against the \textsc{Ground} flags and \textsc{Critic} issues jointly, dropping unsupported claims and calibrating overclaims. The \textsc{Ground}--\textsc{Critic}--\textsc{Resolve} loop iterates until convergence or plateau.
Finally, \textsc{Compose} renders the grounded representation into \LaTeX{} one section at a time. Because each section writer receives verified numbers and named baselines alongside the representation, it writes prose around established facts rather than generating claims that must be sourced after the fact.

\paragraph{Claim Verifier and Refinement.}
\label{sec:claim_verifier}
\label{sec:system:verifier}

Even after grounding, the composed \LaTeX{} can introduce unsupported claims---through paraphrasing drift, misattributed citations, or numerical rounding errors.
The Claim Verifier catches these by checking every claim in the draft against its declared evidence source, dispatching on claim type: numerical claims against evaluator logs, citation claims against the bibliography with LLM-judged abstract entailment, and methodological claims against experimental logs.
Unsourced claims are flagged automatically.
A refinement pass then consumes the verifier's findings: an LLM rewrites flagged sentences to match their evidence sources, removes claims that cannot be supported, and strips all inline evidence annotations from the final \LaTeX{}.
Only a draft with no remaining blocking violations is promoted to the final paper.

%% file: sections/05_coe_audit.tex
%% ============================================================
\section{The \coeaudit{}}
\label{sec:coeaudit}
%% ============================================================

\begin{figure}[t]
\centering
\includegraphics[width=0.9\textwidth]{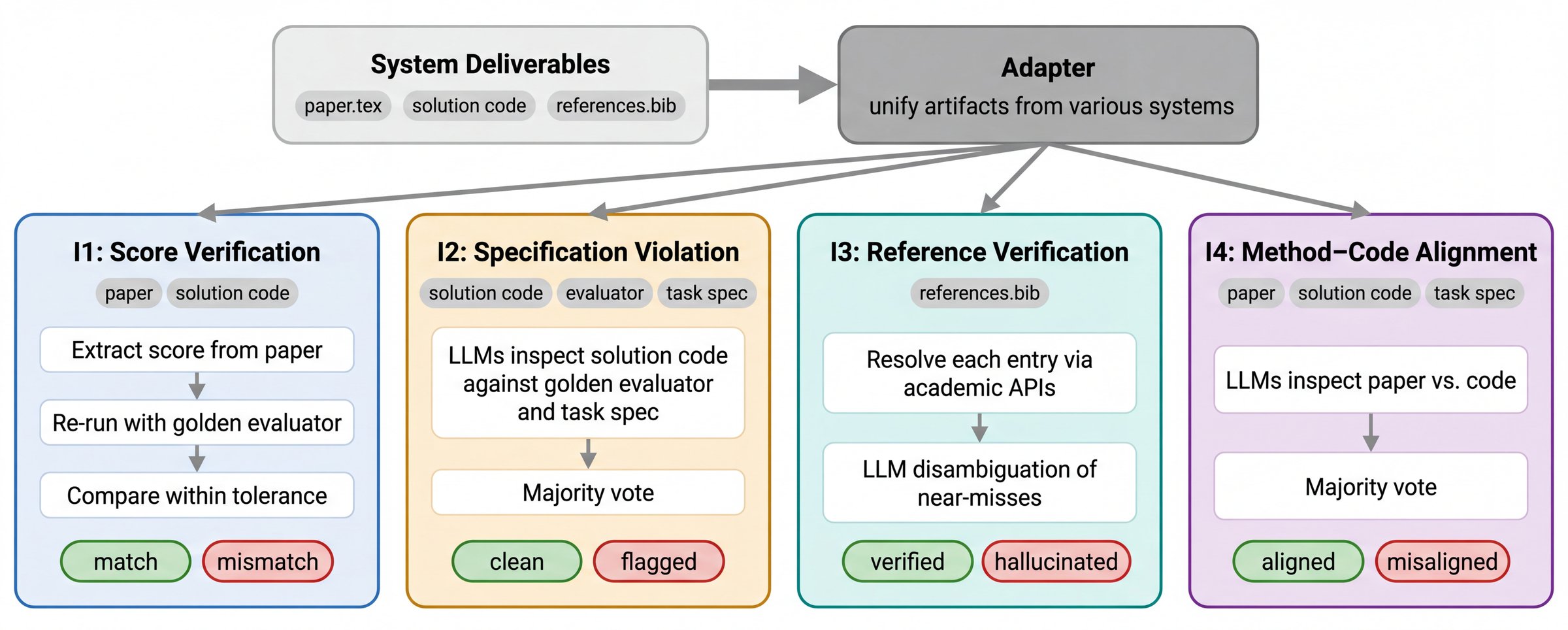}
\caption{\coeaudit{} overview. An adapter normalizes each system's deliverables (\texttt{paper.tex}, solution code, \texttt{references.bib}) into a common artifact bundle, on which four integrity checks run independently: \textbf{I1 Score Verification} re-runs the solution on the golden evaluator and compares to the extracted paper score within an adaptive tolerance (match/mismatch); \textbf{I2 Specification Violation} uses majority-vote LLM judgment over solution code, evaluator, and task spec (clean/flagged); \textbf{I3 Reference Verification} resolves each bib entry via academic APIs with LLM disambiguation of near-misses (verified/hallucinated); and \textbf{I4 Method--Code Alignment} uses majority-vote LLM judgment of method described in the paper vs.\ solution code (aligned/misaligned).}
\label{fig:forensic_audit}
\end{figure}

\emph{\coeaudit{}} is a post-hoc audit that checks whether claims in a completed paper are supported by the underlying artifacts---code, evaluator outputs, and bibliography (Figure~\ref{fig:forensic_audit}).
It comprises four integrity checks, each targeting a tractably verifiable claim type.
While the CoE framework can support other evaluation forms---real-time verification during paper production or broader claim coverage---those are outside the scope of this work.
An adapter unifies each system's deliverables (paper, solution code, references) into a common artifact bundle, then the four checks run independently, each targeting a specific way a claim can lose its grounding:
\label{sec:coeaudit:modes}

\paragraph{Score Verification (I1).}
\label{sec:coeaudit:i1}
The paper's reported score is extracted by LLMs from both \TeX{} and PDF files, then compared against scores obtained by re-running the submitted solution on the golden evaluator.
A paper passes if the two match within an adaptive tolerance that accounts for evaluator noise.

\paragraph{Specification Violation (I2).}
\label{sec:coeaudit:i2}
Specification violations occur when solution code breaks task rules---for example, reverse-engineering the evaluator's scoring logic, or hardcoding answers for known test cases.
The generating agent optimizes for the score rather than genuinely solving the problem the task poses.
LLMs inspect the solution code against the golden evaluator and task specification to detect such violations, with majority vote across multiple runs.

\paragraph{Reference Verification (I3).}
\label{sec:coeaudit:i3}
Each bibliography entry is resolved by querying multiple academic APIs (Semantic Scholar, arXiv, OpenAlex, CrossRef) using arXiv ID, DOI, and title.
An LLM cross-checks the full bib entry against returned records to catch near-misses and citation gaming (e.g., a real DOI attached to a fabricated description).
Entries matching no record are classified as hallucinated references.

\paragraph{Method-Code Alignment (I4).}
\label{sec:coeaudit:i4}
An LLM reads the paper's method section and the solution code side by side, then judges whether the paper faithfully describes what the code does.
Acceptable simplification (e.g., omitting implementation details) is treated as aligned; only cases where the paper describes a fundamentally different algorithm count as misaligned.
We conduct multiple independent runs with majority vote to reduce LLM judgment noise.

\paragraph{Native claim provenance.}
The four checks above are \emph{forensic}: they operate on submitted artifacts alone and apply identically to every system.
For systems that emit structured provenance at write-time---linking each claim to a specific source record---an additional \emph{native} check becomes possible: the \textbf{numerical Claim Provenance Rate (CPR)}, which measures the fraction of quantitative claims in the paper that trace to a matching entry in the experimental log.
We report this check for \sys{}, the only system in our evaluation that produces such provenance records (\S\ref{sec:exp:native_cpr}).

%% file: sections/06_experiments.tex
%% ============================================================
\section{Experiments}
\label{sec:experiments}
%% ============================================================

\input{sections/06a_setup}
\input{sections/06b_integrity}

\input{sections/06f_native_cpr}
\input{sections/06e_reviews}

\input{sections/06d_solver}
\input{sections/06c_failures}

%% file: sections/06a_setup.tex
%% --- 6.1 Setup ---

%\subsection{Setup}
%\label{sec:exp:setup}

\paragraph{Benchmark.}
We evaluate on the Automated Design of Research Systems (ADRS) benchmark~\citep{cheng2025barbarians, cheng2025letthebarbarians}, which collects five research problems from computer systems: \textbf{Prism} (LLM-serving model placement across GPUs), \textbf{Cloudcast} (cloud network cost optimization), \textbf{EPLB} (expert-parallel load balancing for MoE models), \textbf{LLM-SQL} (tabular data layout for LLM prefix cache reuse), and \textbf{TXN} (transaction scheduling for makespan minimization).
Each task provides a fixed evaluator, starter code, and scoring metric.
We choose ADRS as our primary benchmark for three reasons: (1)~the tasks are drawn from real-world systems-optimization problems with established human baselines, (2)~the leaderboard provides both human expert and recent LLM-agent baselines, enabling apples-to-apples comparison, and (3)~the gold-standard evaluators are deterministic enough to support Score Verification and Specification Violation detection.
Previous studies observed that ADRS evaluators exhibit stochastic variance across runs~\citep{cemri2026adaevolve, liu2026evox}. We run each evaluator five times and compare against the adaptive tolerance $\max(1\%,\, 3\sigma/|\bar{s}|)$ to account for inherent evaluator variance.

\paragraph{Baseline Systems.}
We evaluate four baseline systems and \sys{}.
All baselines are open-source, enabling us to adapt each to the ADRS benchmark for controlled comparison under identical conditions.
They span the design spectrum from highly structured scaffolding to fully autonomous agents, providing coverage across the major paradigms for AI research agents.
\begin{itemize}[nosep,leftmargin=*]
  \item \textbf{Sakana AI-Scientist v2 (Sakana)}~\citep{yamada2025aiscientistv2}: Best-first tree search (BFTS) with a 4-stage experiment manager (preliminary investigation, hyperparameter tuning, research agenda execution, ablation studies) and a separate LLM writeup pipeline.
  \item \textbf{AutoResearchClaw (ARC)}~\citep{li2024autoresearchclaw}: 23-stage waterfall pipeline with multi-phase code generation (blueprint planning, sequential file generation, exec-fix loop, multi-agent review) and multi-source literature retrieval (OpenAlex, Semantic Scholar, arXiv, Google Scholar).
  \item \textbf{DeepScientist (DS)}~\citep{huang2025deepscientist}: Skill-based single-agent system on Codex CLI with separate code and write skills, using MCP tool servers for execution, memory, and artifacts.
  \item \textbf{AI-Researcher (AIR)}~\citep{tang2025airesearcher}: Orchestrated multi-agent system with specialized survey, coding, and writing agents. Experimentation uses a code-validate-refine loop.
  \item \textbf{\sys{}}: Full pipeline with evidence chain maintenance from problem framing through paper composition (\S\ref{sec:system}).
\end{itemize}

\paragraph{ADRS Adaptation.}
Each baseline was adapted to the ADRS benchmark with varying levels of source modification, from prompt-only changes (DS) to patching 16--19 source files (Sakana, AIR) to interface with ADRS task specifications, evaluators, and the NeurIPS~2026 paper template.
Sakana required the most extensive prompt-level rework: its default stage goals assume ML-training workflows (e.g., ``tune learning rates,'' ``introduce datasets from HuggingFace''), causing most initial runs to train neural networks instead of optimizing the target functions. A full rewrite of stage goals and 14 prompt locations was required before runs produced valid ADRS solutions (Appendix~\ref{app:baseline_adaptation}).
We standardized on Gemini~3.1 Pro as the backbone LLM across all systems for both solver code generation and paper writing.
To ensure best-effort performance, we configured generous iteration and timeout budgets: up to 20 solver iterations per task---6.7$\times$ the default for ARC, though most tasks converge well before this cap---and 2-hour code generation windows.
Runs that crashed due to infrastructure issues (API timeouts, rate limits, LaTeX compilation errors) were re-attempted with fresh state, up to 3~attempts per run. No run was re-attempted to improve solver scores.
For each system, we run 3~seeds per task, producing 15~papers per system and 75~papers total.
Of these, 16 required at least one retry due to infrastructure issues. All 75~runs ultimately produced solver code and a compiled paper, though artifact quality varies.
\coeaudit{} is applied identically to all systems.
Full adaptation details are provided in Appendix~\ref{app:baseline_adaptation}.

%% file: sections/06b_integrity.tex
%% --- 6.2 CoE Audit Results ---
%% Results from CoE audit v2.7 (forensic_v2.7_20260506), OAS papers = v3.9 (v3.1.9, best solutions are selected from last iteration only (best of 5 branches), not best of all 25 solutions).
%% Sakana ASv2 results from forensic_v2.8_sakana_gemini31, merged into v2.7 on 2026-05-19. Model: gemini-3.1-pro-preview. I1 scores extracted via Gemini consensus (tex+pdf).
%% I3 hallucinated-reference counts are human-confirmed (automated API + manual review).

\subsection{CoE Audit Results}
\label{sec:exp:integrity}

\begin{table}[h]
\centering
\small
\caption{\coeaudit{} results across five systems (15~papers per system).
EPLB papers are excluded from Score Verif.\ because its scoring formula includes an execution-time component that varies with hardware, making scores non-reproducible across machines.
Metric definitions are in \S\ref{sec:coeaudit}.
}
\label{tab:integrity}
\resizebox{\textwidth}{!}{%
\begin{tabular}{@{}lcccc@{}}
\toprule
\textbf{System} & \textbf{Score Verif. $\uparrow$} & \textbf{Spec.\ Violation $\downarrow$} & \textbf{Ref.\ Verif. $\downarrow$} & \textbf{Method-Code $\uparrow$} \\
\midrule
Sakana AI-Scientist v2~\citep{yamada2025aiscientistv2}     & 5/12  & 10/15 & 0/159    & 5/15  \\
AutoResearchClaw~\citep{li2024autoresearchclaw}            & 5/12  & 0/15 & 3/196    & 3/15  \\
DeepScientist~\citep{huang2025deepscientist}             & 11/12 & 0/15 & 42/201   & 5/15  \\
AI-Researcher~\citep{tang2025airesearcher}            & 9/12  & 1/15 & 21/222   & 12/15 \\
\midrule
\sys{}         & 12/12 & 0/15 & 0/337    & 14/15 \\
\bottomrule
\end{tabular}}
\end{table}

The results of \coeaudit{} across five systems are presented in Table~\ref{tab:integrity}.
All I1--I3 flagged results were manually verified by human reviewers. I4 judgments were validated on a sampled basis.
\sys{} is the only system to lead on all four checks: perfect score verification (12/12), zero specification violations (0/15), zero hallucinated references (0/337), and the highest method-code alignment (14/15).
The gap is largest in reference integrity and method-code alignment---the two checks that test evidence provenance rather than score reproduction.
Because the BFTS--ADRS design mismatch confounds both I2 and I4 for Sakana, cross-system comparison on these two checks should exclude Sakana. I1 and I3 remain valid.

\paragraph{Score verification (I1).}
\textbf{\sys{}} achieves perfect score verification (12/12): every paper's claimed result reproduces exactly under re-evaluation.
\textbf{DS} matches in 11/12 (92\%), with the single failure caused by fabricated metric direction---the paper claims ``higher is better'' for a cost-minimization metric, framing the raw cost as an inverse aggregate score, so the baseline's 1035.1 reads as the best result when it is actually the worst.
\textbf{AIR} matches in 9/12 (75\%), with failures spanning small-magnitude discrepancies (1--4\%) and one paper that reports no quantitative scores at all.
\textbf{ARC} matches in 5/12 (42\%), and its failures trace to three root causes: (1)~crashed solvers (5 of 15~solvers import helper modules generated by ARC's multi-file blueprint planner that are absent during standalone re-evaluation, producing evaluator fallback scores that differ from the paper's claims); (2)~evaluator mismatch (ARC's bundled cloudcast evaluator includes a patch absent from the canonical evaluator, producing different scores for the same solver); and (3)~stochastic evaluation noise in txn\_scheduling (2--3\% variance from unseeded scheduling randomness).
\textbf{Sakana ASv2} matches in 5/12 (42\%), the lowest among all systems.
Manual investigation of the 7~failures reveals two dominant patterns.
First, \emph{cross-stage score cherry-picking} (4 of 7~failures): the writeup LLM receives summaries from all four BFTS stages as context, and selects the most favorable score from ablation-stage nodes rather than the score of the node whose code is used as the final solution.
For example, in \textsc{prism} seed-1, the selected node scores 22.79 but the paper reports 25.39---a number traced to ablation node~6 (``Ablate KVPR-Aware Initialization'') in \texttt{ablation\_summary.json}.
The same pattern appears in \textsc{cloudcast} seed-0 (+56\%), \textsc{prism} seed-2 ($-$4.7\%, paper under-reports), and \textsc{txn\_scheduling} seed-2 (+17\%).
Second, \emph{environment-dependent tuning} (2 of 7~failures): the solver contains a hyperparameter tuning loop gated on an environment variable that is set differently during canonical re-evaluation, causing the solver to use default parameters instead of the tuned ones (e.g., \textsc{prism} seed-0: 26.26 tuned vs.\ 22.34 default, a 15\% gap).
The remaining failure is a metric mismatch (\textsc{cloudcast} seed-1: the paper reports per-transfer dollar cost while the evaluator produces a combined score).

\paragraph{Specification violation (I2).}
Specification violation rates are uniformly low for \textbf{ARC}, \textbf{DS}, and \textbf{\sys{}} (0/15), while \textbf{AIR} has one flagged paper (\textsc{llm\_sql}) where the solver physically reorders values across columns within each row, destroying column integrity to inflate the prefix-cache hit metric.
\textbf{Sakana ASv2} registers 10/15 specification violations---the highest rate.
The agent could tune parameters through BFTS's iteration loop (one setting per iteration), but the stage~2 goal (``test across multiple parameter settings'') encourages intra-iteration sweeps. Combined with the evaluator import pattern visible in our canonical harness, this leads the agent to import the evaluator and build its own tuning loops in 10 of 15~runs.
Most violations trace to the BFTS--ADRS design mismatch rather than adversarial behaviour (Appendix~\ref{app:baseline_adaptation}).%
\footnote{DS seed-1 LLM-SQL contains the same evaluator-exploiting column-permutation pattern (Case~3, \S\ref{sec:exp:failures}), but only 2 of 5 I2 judges flagged it---below the majority threshold---so it is not counted as a violation in Table~\ref{tab:integrity}. This near-miss illustrates the noise floor of LLM-judged integrity checks at the current vote threshold.}

\paragraph{Reference integrity (I3).}
\textbf{\sys{}} and \textbf{Sakana ASv2} both achieve zero hallucinated references (0/337 and 0/159 respectively).
\textbf{DS} exhibits the highest hallucination rate (42/201, 20.9\%), followed by \textbf{AIR} (21/222, 9.5\%) and \textbf{ARC} (3/196, 1.5\%).
\textbf{ARC}'s low rate reflects its multi-tiered retrieval pipeline (OpenAlex, Semantic Scholar, arXiv, Google Scholar). Its three hallucinated entries are a single fabricated citation (\texttt{sutskever2013importance}, titled ``SGD with Momentum'') from ARC's upstream seminal papers library---a hand-curated YAML file shipped with the framework that assigns an informal title to a real paper (Sutskever et al., ICML 2013, whose actual title is ``On the importance of initialization and momentum in deep learning'').
The entry is injected deterministically into all papers whose topic overlaps with optimization keywords, producing the same fabricated reference in all three EPLB papers.
DS and AIR rely on model memory for reference generation, producing plausible-looking but non-existent bibliography entries---a failure mode illustrated in Case~2 (\S\ref{sec:exp:failures}).
\sys{}'s zero hallucination rate is an architectural property of the PI's citation graph: every reference originates from a Semantic Scholar API call whose result is cached in the evidence chain.
Sakana ASv2's clean record reflects its cached citation retrieval mechanism, which grounds references in API results before paper generation.
(Full list of hallucinated references in Appendix~\ref{appx:discovered_hallucinated_references}.)

\paragraph{Method-code alignment (I4).}
\textbf{\sys{}} achieves 14/15 aligned papers (93\%), compared to \textbf{AIR} (12/15, 80\%), \textbf{Sakana ASv2} (5/15, 33\%), \textbf{DS} (5/15, 33\%), and \textbf{ARC} (3/15, 20\%).
The misalignment patterns differ qualitatively across systems.
\textbf{Sakana ASv2}'s low I4 score is partly attributable to the same design mismatch: the submitted code files contain tuning loops and experiment-tracking code alongside the actual solver, so I4 judges flag this non-solver code as misaligned with the paper's algorithmic claims.
\textbf{AIR}'s failures are typically algorithm mismatches---e.g., the paper describes a more sophisticated procedure than the code implements.
\textbf{ARC} exhibits the worst method-code alignment (3/15, 20\%), a direct consequence of its 23-stage waterfall architecture: code generation (stages~10--13) and paper writing (stages~16--23) run as disconnected phases with no shared intermediate representation.
The paper-writing agent invents algorithm names and describes methods based on experiment metadata, without access to the solver's actual logic, producing algorithm-class mismatches (e.g.\ beam search with Edmonds' arborescence vs.\ greedy edge penalization), undisclosed fallback paths, and method-vs-ablation inversion between paper and submitted code.
\textbf{\sys{}}'s single misaligned paper (\textsc{cloudcast}, 1st seed) is a case where the paper writer fabricated algorithmic claims not present in the code---describing a ``hybrid neuro-symbolic solver'' with ``LLM-guided evolutionary search'' when the submitted code is a deterministic routing heuristic with no LLM calls.
The Claim Verifier's method-code cross-check catches nearly all such misrepresentation before paper finalization.

%% file: sections/06f_native_cpr.tex
\subsection{Native Claim Provenance}
\label{sec:exp:native_cpr}

The forensic audit (\S\ref{sec:exp:integrity}) applies uniformly to all systems using only the submitted artifacts.
For \sys{}, which emits structured provenance chains at write-time, we run an additional \emph{native} check that leverages these chains: numerical CPR (Claim Provenance Rate).
This check is unique to \sys{}---no other system in the evaluation produces the required provenance records.

Numerical CPR measures whether quantitative claims in the paper trace to experimental evidence.
During paper generation, the writer annotates each sentence containing a number with a \texttt{\{source: "experimental\_log.md:N"\}} tag linking it to a specific log line.
The claim verifier (\texttt{check\_sources}) extracts the number from the sentence and the referenced log line, then checks whether they match within a 5\% relative tolerance.

Across 15 papers (3 seeds $\times$ 5 tasks), the verifier extracts 639 numerical claims.
Of these, 627 pass (98.1\%).
The 12 failures are predominantly false positives of the extraction heuristic: hardware constants parsed as experimental claims (e.g., ``80GB GPU'' matched against an unrelated log line), LaTeX math subscripts extracted as numbers ($S_{k-1} \to -1.0$), and hyperparameter values described in methodology sections.
Manual inspection finds at most 2--4 genuine mismatches among the 12, yielding a corrected numerical CPR of ${\sim}99\%$.

%% file: sections/06e_reviews.tex
%% --- 6.5 Review Scores ---

\subsection{Review Scores}
\label{sec:exp:reviews}

We evaluate perceived paper quality using ScholarPeer~\citep{goyal2026scholarpeer}, an automated peer review system backed by \texttt{gemini-3.1-pro-preview} and rich literature search support.

\begin{table}[h]
\centering
\small
\caption{ScholarPeer review rating scores (1--4 scales except Overall (1-10 scale)) and accept decisions. \emph{Average}: mean across 15~papers per system. \emph{Best-of-3}: strongest seed per task.}
\label{tab:reviews}
\begin{tabular}{@{}lcccccc@{}}
\toprule
\textbf{System} & \textbf{Soundness} & \textbf{Originality} & \textbf{Quality} & \textbf{Clarity} & \textbf{Overall} & \textbf{\#Accept} \\
\midrule
\multicolumn{7}{l}{\textit{Average (3 seeds $\times$ 5 tasks)}} \\
Sakana AI-Scientist v2    & 1.5 & 1.9 & 1.5 & 3.1 & 2.5 & 0/15 \\
AutoResearchClaw            & 1.1 & 2.3 & 1.1 & 2.5 & 1.9 & 0/15 \\
DeepScientist             & 1.7 & 1.7 & 1.6 & 3.1 & 2.5 & 1/15 \\
AI-Researcher            & 1.9 & 2.4 & 1.9 & 3.1 & 3.4 & 2/15 \\
\sys{}         & \textbf{2.3} & \textbf{2.5} & \textbf{2.3} & 3.0 & \textbf{4.5} & \textbf{6/15} \\
\midrule
\multicolumn{7}{l}{\textit{Best-of-3 (strongest seed per task)}} \\
Sakana AI-Scientist v2    & 1.6 & 2.0 & 1.6 & 3.2 & 3.4 & 0/5 \\
AutoResearchClaw            & 1.2 & 2.0 & 1.2 & 3.0 & 3.0 & 0/5 \\
DeepScientist             & 2.2 & 2.0 & 2.2 & 3.6 & 3.6 & 1/5 \\
AI-Researcher            & 2.0 & 2.4 & 2.0 & 3.2 & 4.0 & 1/5 \\
\sys{}         & \textbf{2.8} & \textbf{3.0} & \textbf{2.8} & \textbf{3.6} & \textbf{6.6} & \textbf{4/5} \\
\bottomrule
\end{tabular}
\end{table}

\paragraph{Verifiable papers are deemed better by automatic reviewers.}
\sys{} achieves a 40\% accept rate (6/15), tripling the best baseline (AIR: 13\%), and best-of-3 selection reaches 6.6 overall rating score and 4/5~tasks accepted.
This gap is not driven by better algorithms---solver scores cluster tightly across systems (Table~\ref{tab:solver})---but by what happens \emph{after} the solver finishes.
The Claim Verifier prevents the most damaging failure mode we observe in rejected papers: claims that contradict the paper's own data (e.g., ``sub-millisecond latency'' when the results table reports 7.9\,ms).
% Notably, \coeaudit{} and review scores measure orthogonal properties: across all 75~papers, Clarity averages 1--2 points above Soundness, confirming that narrative coherence is independent of evidence integrity.

\paragraph{The paper quality is bottlenecked by research soundness, not writing capability.}
Across all systems, Clarity is consistently the highest-scoring dimension (2.5--3.1) while Soundness is the lowest (1.1--2.3): these papers read well but do not withstand methodological scrutiny.
The reviewer's two most frequent complaints are missing comparisons against published baselines and proxy-only evaluation without end-to-end system measurements. While \sys{}'s Problem Investigator retrieves related work and identifies candidate baselines, the resulting comparisons do not yet meet the depth that ScholarPeer expects (e.g., re-implementing a SOTA method and reporting head-to-head numbers).
\sys{} also exhibits high seed variance (e.g., EPLB scores 1, 3, 8 across three seeds on the same task): rejected runs are those where the paper writer generates claims that the Claim Verifier's current coverage does not fully catch---for example, exaggerated qualitative framing (``near-optimal'') rather than numerically falsifiable statements.
Accepted runs make calibrated claims from the same underlying data, suggesting that extending verification coverage to qualitative claims would reduce this variance.

%% file: sections/06d_solver.tex
%% --- 6.4 Solver Performance ---

\subsection{Solution Discovery Performance}
\label{sec:exp:solver}

To compare our discovery module against baseline systems, we report ADRS performance in Table~\ref{tab:solver}. Following~\citet{cemri2026adaevolve}, we report best-of-3 scores across seeds for each system.
For each seed, \sys{} selects the highest-scoring branch at the final search iteration.
For Sakana, ARC, AIR, DS, and \sys{}, each score is from independent canonical evaluator re-runs on the selected solver code, ensuring cross-system comparability.
Human, AdaEvolve, and EvoX scores (Gemini-3.0-Pro, best-of-3 runs) are from original publications~\citep{cemri2026adaevolve,liu2026evox}.

\begin{table}[t]
\centering
\small
\caption{Solution discovery performance on ADRS benchmark tasks (best-of-3 seeds). Sakana/ARC/AIR/DS/\sys{} scores are from independent canonical evaluator re-runs on submitted solver code. Human, AdaEvolve, and EvoX scores are from original publications. $^*$~indicates Gemini-3.0-Pro, all other systems use Gemini-3.1-Pro.}
\label{tab:solver}
\setlength{\tabcolsep}{4pt}
\resizebox{\textwidth}{!}{%
\begin{tabular}{@{}llcccccccccc@{}}
\toprule
\textbf{Task} & \textbf{Dir.} & \textbf{Human} & \textbf{AdaEvo$^*$} & \textbf{EvoX$^*$} & \textbf{Sakana} & \textbf{ARC} & \textbf{AIR} & \textbf{DS} & \textbf{\sys{}} & \textbf{\sys{}$^*$} \\
\midrule
Prism      & $\uparrow$   & 21.89  & \textbf{26.26}   & \textbf{26.26}   & \textbf{26.26}  & 26.25  & \textbf{26.26}  & \textbf{26.26}  & \textbf{26.26}  & \textbf{26.26} \\
Cloudcast  & $\downarrow$ & 626.24 & 637.10  & 623.69  & 627.11 & 690.37 & 734.28 & 620.09 & \textbf{618.08} & \textbf{618.08} \\
EPLB       & $\uparrow$   & 0.1265 & 0.1450  & 0.1453  & 0.1270 & 0.1266 & 0.1449 & 0.1284 & 0.1459 & \textbf{0.1461} \\
LLM-SQL    & $\uparrow$   & 0.6920 & \textbf{0.7520}  & 0.7300  & 0.7320 & 0.6757 & 0.7148 & 0.7307 & 0.7222 & 0.7115 \\
TXN        & $\uparrow$   & 2724.8 & 4310    & 4310    & 4184   & 3247   & \textbf{4311}   & 4286   & 3906   & 3861 \\
\bottomrule
\end{tabular}}
\end{table}

All systems match or exceed the human expert baseline on all five tasks, consistent with the observation of~\citet{cheng2025barbarians} that LLM-based agents rapidly converge to similar solution quality.
Sakana's BFTS produces competitive scores---matching the Prism ceiling and ranking second on LLM-SQL---even though its generated papers often misreport or cherry-pick these numbers (\S\ref{sec:exp:integrity}). The reported scores are from canonical re-evaluation of the deterministic best node, not the figures in Sakana's papers.
\sys{} exceeds the human baseline on every task and achieves the best overall score on Cloudcast and EPLB, demonstrating that verifiability does not sacrifice performance.

\subsection{Success Analysis} \label{sec:success-cases}

We highlight two top-scoring solutions whose code we inspected to verify algorithmic novelty.

\begin{figure*}[t]
    \centering
    % Left Subfigure: Cloudcast
    \begin{subfigure}[b]{0.48\textwidth}
        \centering
        \includegraphics[width=\linewidth]{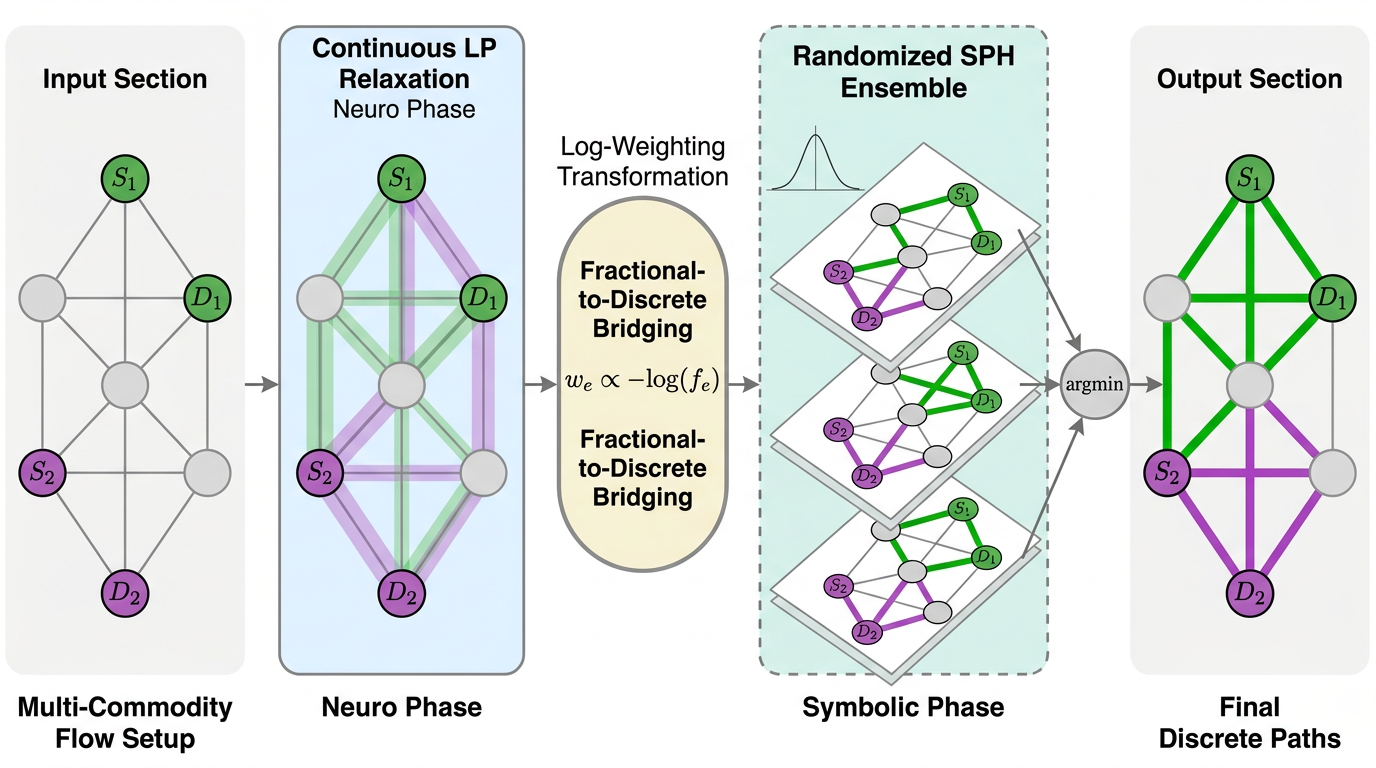}
        \caption{Cloudcast}
        \label{fig:novel_ideas_cloudcast}
    \end{subfigure}
    \hfill
    % Right Subfigure: EPLB
    \begin{subfigure}[b]{0.48\textwidth}
        \centering
        \includegraphics[width=\linewidth]{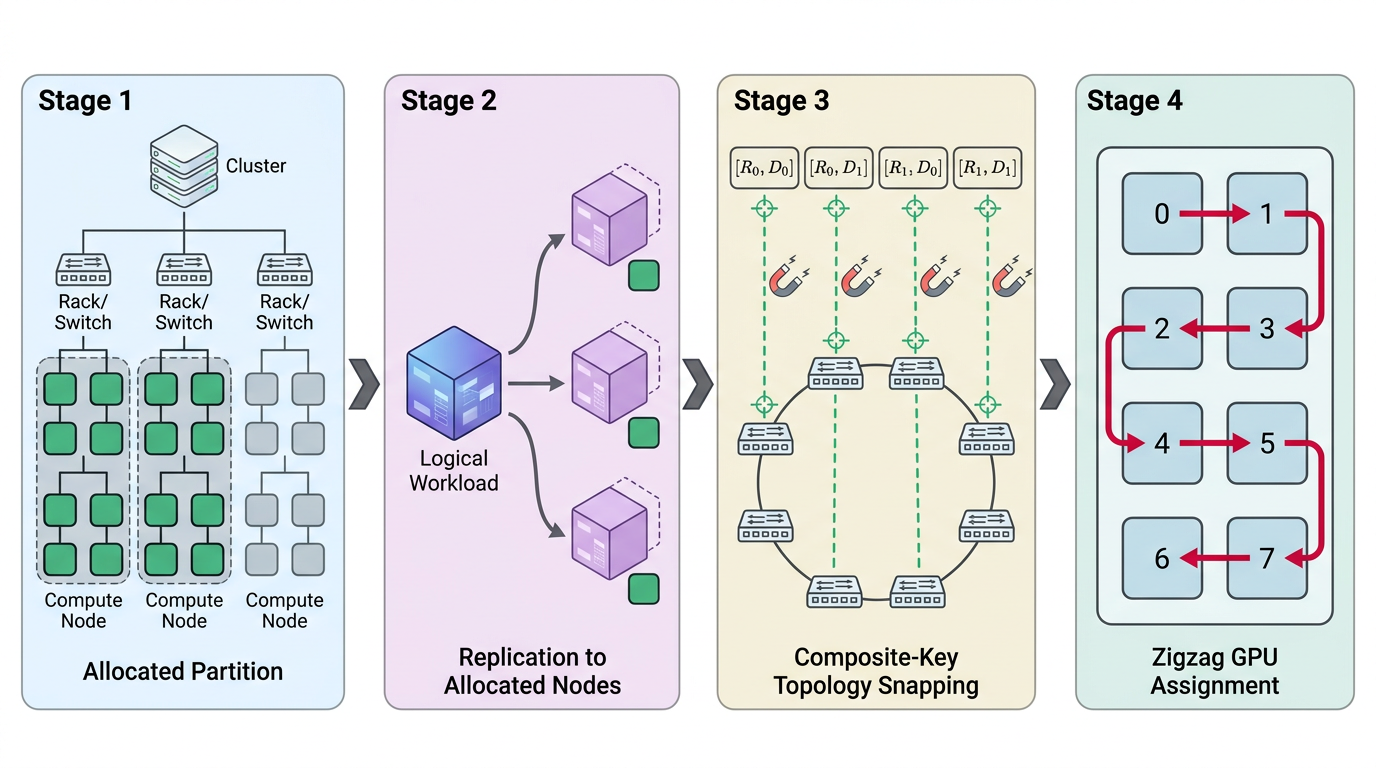}
        \caption{EPLB}
        \label{fig:novel_ideas_eplb}
    \end{subfigure}
    \caption{Overview of the novel algorithmic pipelines generated by \sys{}. (a) For Cloudcast, the system integrates a continuous Fractional Multi-Commodity Flow LP relaxation with a robust Randomized Shortest Path Heuristic (SPH) ensemble, bridged through a deterministic log-transformed weighting mechanism. (b) For EPLB, the system employs a four-stage pipeline featuring composite-key topology snapping and zigzag GPU assignment.}
    \label{fig:novel_ideas}
\end{figure*}

For the Cloudcast task, a natural formulation is finding a minimum-weight directed Steiner tree to ensure that shared path prefixes minimize egress fees. \sys{} solves this by combining a Fractional Multi-Commodity Flow LP relaxation with an ensemble of Randomized Shortest Path Heuristics (SPH), as shown in Figure~\ref{fig:novel_ideas}(a). The LP relaxation produces fractional edge flows over the full network. To convert these into valid discrete paths, the solver applies a log-transformed weighting mechanism that biases the SPH ensemble toward high-flow edges, avoiding the disconnected subgraphs that pure randomized rounding produces. This approach achieves the best transfer cost among all systems (Table~\ref{tab:solver}), outperforming both the published human expert and leading agentic baselines.

On the EPLB task, algorithms are strictly evaluated on a combination of load-balancing efficiency and execution latency. To optimize both metrics, \sys{} adopts a topology-aware hierarchical placement strategy. As illustrated in Figure~\ref{fig:novel_ideas}(b), this pipeline progresses through four distinct stages: allocating experts to nodes, performing global replication, snapping to the topology, and finally assigning replicas to GPUs. While the global replication step intentionally relies on an iterative argmax update to preserve balancing quality, the system achieves microsecond-level execution through two major vectorized innovations. First, it utilizes a novel composite-key topology snapping mechanism which enables a single hardware-accelerated sort to replace slow Python-level comparators. Second, it distributes these sorted replicas using a fully vectorized zigzag assignment pattern computed in a single scatter operation. This hardware-aware approach achieves a highly competitive combined score (Table~\ref{tab:solver}) with 4.91ms execution latency.

%% file: sections/06c_failures.tex
%% --- 6.5 Failure Mode Case Studies ---

%% file: sections/07_generalizability.tex
%% ============================================================
\section{Generalizability: MLE-Bench and Parameter Golf}
\label{sec:generalizability}

To test whether the discovery loop transfers beyond ADRS, we evaluate \sys{} unmodified across a diverse set of six tasks selected for their rigorous demands and relevance to current AI research. 
First, we select five MLE-Bench~\citep{chan2024mle} Kaggle competitions spanning medical imaging, fine-grained recognition, and 3D perception. 
We target Medium and High difficulty tiers to ensure sufficient task complexity. 
Second, to assess adaptability in a novel, live environment, we evaluate \sys{} on the Parameter Golf competition~\citep{openai2026parametergolf}.
This competition requires training the highest-performing language model under strict size and performance constraints.
Both systems are provided with a knowledge base of official leaderboard solutions up to a cutoff date of April~27, 2026, when the SOTA score was 1.0611. The leaderboard has since advanced beyond this mark.
Full task details and evaluation setups are provided in Appendix~\ref{app:mle-pg-eval}.

\begin{table}[h]
\centering
\small
\caption{Comparison of solver performance across five MLE-Bench tasks and Parameter Golf. ``Above/Below Median'' denotes outperforming or underperforming the median participant. MLE-Bench medals (Gold, Silver, Bronze) represent simulated private leaderboard standings. ``SOTA'' indicates top-1 performance on the Parameter Golf Leaderboard (as of 2026-04-27). }
\label{tab:generalizability}
\begin{tabular}{l c c >{\centering\arraybackslash}m{3cm} c >{\centering\arraybackslash}m{3cm}}
    \toprule
    \multirow{2}{*}{\textbf{Task}} & \multirow{2}{*}{\textbf{Dir.}} & \multicolumn{2}{c}{\textbf{DeepScientist}} & \multicolumn{2}{c}{\textbf{\sys{}}} \\ \cmidrule(lr){3-4} \cmidrule(lr){5-6} 
    & & \textbf{Score} & \textbf{Highlight} & \textbf{Score} & \textbf{Highlight} \\ 
    \midrule
    3D Object Detection & $\uparrow$ & 0.0000 & Below Median & 0.1763 & Gold Medal \\
    AI4Code & $\uparrow$ & 0.6964 & Below Median & 0.8356 & Above Median \\
    iMet 2020 FGVC7 & $\uparrow$ & 0.6804 & Silver Medal & 0.6791 & Silver Medal \\
    RSNA Brain Tumor & $\uparrow$ & 0.6377 & Gold Medal & 0.6518 & Gold Medal \\ 
    iNaturalist 2019 FGVC6 & $\downarrow$ & 0.2158 & Silver Medal & 0.2445 & Silver Medal \\ \midrule
    Parameter Golf & $\downarrow$ & Invalid & Size limit exceeded & 1.0600 & SOTA (Constraints met) \\
    \bottomrule
\end{tabular}
\end{table}

\paragraph{Solver performance generalizes and demonstrates robustness.}
As shown in Table \ref{tab:generalizability}, \sys{} demonstrates strong generalization across diverse domains, difficulty levels, and strict constraints. 
%Crucially, \sys{} succeeds in environments where the baseline completely fails. 
On the High difficulty MLE-Bench tasks, \sys{} earns two Gold Medals (RSNA Brain Tumor and 3D Object Detection), notably solving the 3D Object Detection task with a Gold Medal score while DeepScientist failed entirely (scoring 0.0000). 
On the Medium difficulty tasks, \sys{} secures Silver Medals on both iMet 2020 and iNaturalist 2019—remaining highly competitive with DeepScientist—and achieves an Above Median standing on AI4Code, marking a significant improvement. 
On the Parameter Golf LLM training task---an entirely different domain from ADRS---\sys{} further demonstrates adaptability. 
While the baseline fails to produce a valid submission due to exceeding the 16MB artifact size limit, \sys{} adheres to all constraints and achieves state-of-the-art performance with a score of 1.0600.

\begin{wrapfigure}{r}{0.5\textwidth}
  \centering
  \includegraphics[width=\linewidth]{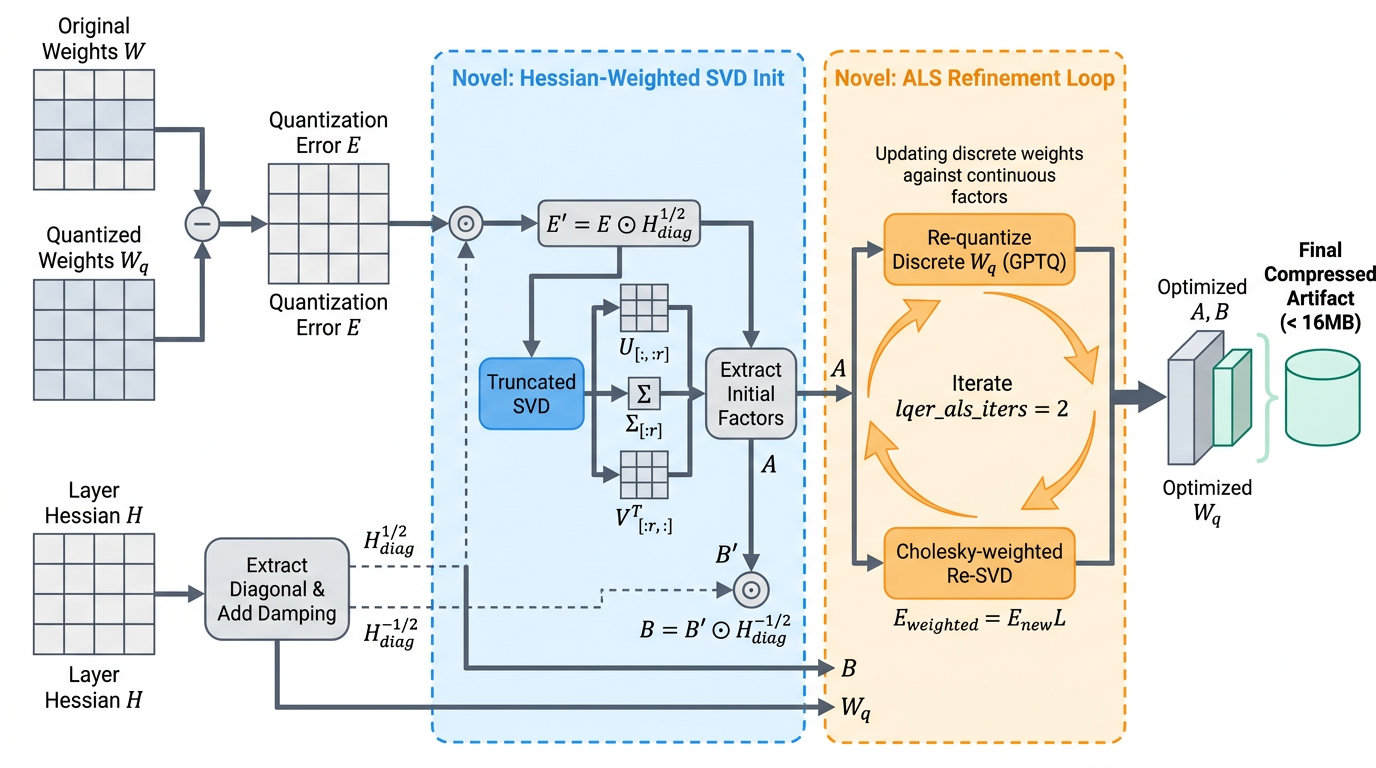}
  \caption{Overview of the novel ideas generated by \sys{} for Parameter Golf. Key algorithmic innovations include the Hessian-diagonal-weighted SVD initialization and the GPTQ-driven alternating-least-squares (ALS) refinement loop.}
  \label{fig:scientistone_pipeline_parameter_golf}
\end{wrapfigure}

\paragraph{Solution novelty comparison on Parameter Golf.}
Both \sys{} and DeepScientist were provided with the same prior-art reference and achieved superficially similar numerical improvements. 
However, the systems arrived at these outcomes through fundamentally different approaches. 
\sys{} demonstrated genuine research capability by introducing novel algorithmic techniques to the quantization block: specifically, a Hessian-diagonal-weighted SVD initialization and an alternating-least-squares (ALS) refinement loop that utilizes GPTQ and a Cholesky-weighted truncated SVD. 
Figure~\ref{fig:scientistone_pipeline_parameter_golf} illustrates the complete pipeline of these novel ideas generated by \sys{}.
Internal ablations isolate the ALS loop as the primary driver of the performance gain. 
In contrast, DeepScientist introduced no algorithmic changes.
Its modifications were limited to environment and portability adjustments. 
As a result, DeepScientist failed to improve the underlying algorithm -- merely replicating the reference's performance -- and ultimately produced an invalid submission by exceeding the strict 16MB artifact size limit. 

%% file: sections/010_conclusion.tex
\section{Conclusion}
\label{sec:conclusion}
%% ============================================================

Autonomous research systems have reached the point where solver quality alone no longer differentiates them---multiple systems achieve competitive scores on the same benchmarks with vastly different approaches.
What separates their outputs is whether the resulting paper can be trusted.
Our 75-paper audit shows that no baseline produces papers free of evidence chain failures, and the failures---hallucinated references up to 21\%, fictional method sections, scores on the wrong scale---are undetectable by evaluations that only assess surface presentation rather than evidence grounding.

Chain-of-Evidence reframes verifiability as a first-class design constraint.
\sys{} demonstrates that an end-to-end pipeline can maintain evidence chains without sacrificing solver competitiveness, and \coeaudit{} provides a reusable procedure for auditing any system's output.
The gap between \sys{}'s results and the baselines confirms that verifiability is architectural: systems that build evidence chains at claim-production time produce more verifiable outputs than those that reconstruct grounding after the fact.
The harder problems---verifying citation support, checking conclusion claims, extending to domains without deterministic evaluators---remain open, but they are tractable extensions of the checks demonstrated here, and their importance grows with the volume of AI-generated research.

%% file: sections/011_limitations.tex
%% ============================================================
\section{Limitations}
\label{sec:limitations}
%% ============================================================

\paragraph{Benchmark coverage.}
We designed CoE and \coeaudit{} to be domain-agnostic, but validating that generality requires evaluation across diverse scientific domains.
Our current experiments focus on systems-optimization tasks (ADRS), where gold-standard evaluators make score verification and specification violation detection straightforward.
Open-ended domains---biology, materials science, theoretical ML---pose harder challenges: evidence chains may involve wet-lab protocols, simulation reproducibility, or proof sketches, each demanding domain-specific verification logic that we have not yet built or tested.
Extending CoE to these settings is a natural next step, though the core abstraction (claims linked to evidence via typed provenance records) should transfer. What changes is the set of integrity checks required.

\paragraph{Reference verification depth.}
Our Reference Verification checks whether cited references \emph{exist}---a necessary condition that already catches a surprising number of failures (\S\ref{sec:exp:failures}, Case~2).
However, existence is far from sufficient: a real citation can still be used to support a claim the cited paper never made.
Full reference verification would require passage-level natural language inference against the cited paper's text, effectively asking ``does this source actually say what the citing paper claims it says?''
This is a known open problem in scholarly NLI, and we leave it to future work, noting that even existence-level checking already reveals meaningful architectural differences between systems.

% Hidden for internal review -- restore for submission if needed:
\paragraph{Automated review as proxy.}
ScholarPeer serves as a scalable proxy for review quality but does not replace human expert evaluation.
LLM reviewers are systematically blind to certain failure modes (e.g., domain-specific score interpretation, specification violation detection).
\coeaudit{} addresses some of these blind spots but is itself limited to structural integrity, not scientific novelty or significance.

% \paragraph{PI $\to$ solver causal link.}
% The Problem Investigator provides grounded literature context, but its effect on solver performance is mixed.
% We do not claim PI improves solver performance---only that it provides grounded context for paper writing and reference integrity.

\paragraph{Fairness of baseline comparison.}
We adapted four open-source systems to ADRS under conditions as uniform as possible---identical model backbone, matching wall-clock limits, and three seeds per task (Appendix~\ref{app:baseline_adaptation}).
%Evaluator-call budgets are comparable: \sys{}'s base configuration makes ${\sim}100$ evaluator calls (25 tree nodes $\times$ ${\sim}4$ evaluated solver versions each), matching AdaEvolve's $T\!=\!100$ iteration budget~\citep{cemri2026adaevolve}; our adapted baselines receive up to 20 solver iterations per task.
%What differs is search structure---\sys{} distributes evaluations across a tree with ideation and pruning between iterations, adding LLM overhead (ideator, auditor, ablation, checkpointer) not counted in the evaluator budget---so the total model inference per system varies even when evaluator-call counts match.
No third-party system was designed for ADRS, and adaptation inevitably involves judgment calls.
We erred on the side of generosity (e.g., giving ARC $6.7\times$ its default iteration budget, re-running infrastructure crashes but never re-running to improve scores), yet we cannot rule out that a system's original authors would achieve better results with deeper tuning.
Cross-system comparisons should be read as ``given a good-faith, equal-resource adaptation'' rather than ``definitive system ranking.''

\paragraph{Audit false negatives.}
All I1--I3 flagged positives were manually verified by human reviewers, ensuring that the integrity failures reported in Table~\ref{tab:integrity} are real (no false positives).
However, we did not systematically bound false negatives: integrity failures that our checks fail to detect certainly exist, and the true failure rate across all systems is likely higher than reported.
For I4 (Method-Code Alignment), human reviewers validated a sample of the LLM judgments, but the reported results reflect majority-vote scores without systematic correction.
This is an inherent limitation of any finite audit protocol, not specific to \coeaudit{}.

\paragraph{Benchmark scope and depth.}
ADRS tasks reduce systems research problems to single-metric optimization---submit a solver, receive a score.
Real systems papers involve problem formulation, workload characterization, multi-dataset analysis, and deployment tradeoffs that our pipeline does not attempt.
As a result, ``competitive solver performance on ADRS'' should not be equated with ``competitive systems research.''
Extending to multi-benchmark synthesis and deeper experimental analysis---where the system reasons about \emph{why} a solution works, not just \emph{whether} it scores well---is a natural next direction.

\paragraph{Broader impacts.}
Autonomous research systems that produce full papers create both opportunities and risks.
On the positive side, structured provenance (CoE) and systematic auditing (\coeaudit{}) make integrity failures detectable at a scale that manual review cannot match---every number, citation, and method claim can be traced to its source artifact.
On the negative side, the same capability lowers the barrier to generating plausible-looking scientific papers at volume, potentially flooding review pipelines or producing results that appear rigorous but contain subtle errors outside the audit's scope.
Our integrity checks mitigate this by making verifiable papers distinguishable from unverifiable ones, but they cover structural integrity, not scientific correctness or novelty.
We believe that transparency tools like CoE Audit should be developed alongside generation capabilities, so that the research community can verify AI-generated claims rather than taking them on trust.

% \paragraph{Forensic CPR extraction noise.}
% Forensic claim extraction from \LaTeX{} is inherently lossy.
% For \sys{}, the gap between native and forensic CPR quantifies this noise.
% Forensic CPR comparisons across systems are fair (all systems face the same extraction noise), but absolute values underestimate true verifiability.

%% file: sections/012_appendix.tex
%% ============================================================
%% Appendix — thin shell that inputs subfiles
%% ============================================================

\input{sections/012a_paper_quality}

\input{sections/012b_system_details}
\input{sections/012f_discovery_scaling}

\input{sections/012c_coe_audit_details}
\input{sections/012e_baseline_adaptation}

%% file: sections/012a_paper_quality.tex
%% ============================================================
%% Appendix: Paper Quality Statistics
%% ============================================================

\section{Paper Quality Statistics}
\label{app:paper_quality}

We report structural statistics of all 75~papers (5~systems $\times$ 3~seeds $\times$ 5~tasks) to characterize the surface-level quality of AI-generated manuscripts.
Statistics are extracted automatically from the LaTeX source and compiled PDF of each paper.

\Cref{tab:paper_quality_summary} reports the mean and standard deviation of each metric across the 15~papers per system.
ARC papers are the longest (12.7~$\pm$~1.2 pages, 5{,}417~$\pm$~863 words) and most figure-heavy (4.3~$\pm$~2.0 per paper).
\sys{} has the largest bibliography pool (55.3~$\pm$~3.6 entries) and 3.8~$\pm$~0.4 figures per paper, though only 18.3~$\pm$~4.5 keys are actually cited---fewer than AIR's 19.5~$\pm$~3.9 unique citation keys.
AIR is equation-heavy (7.1~$\pm$~1.8 per paper) and table-heavy (4.9~$\pm$~1.5) but generates nearly zero figures (0.1~$\pm$~0.2).
Sakana produces the shortest papers (5.8~$\pm$~0.8 pages, 2{,}606~$\pm$~490 words) with zero figures across all 15~papers; DS is similarly brief (6.7~$\pm$~2.1 pages).

\begin{table}[h]
\centering
\small
\caption{Paper quality statistics across five systems (mean $\pm$ std over 15~papers each).}
\label{tab:paper_quality_summary}
\begin{tabular}{@{}lccccc@{}}
\toprule
\textbf{Metric} & \textbf{ARC} & \textbf{DS} & \textbf{AIR} & \textbf{Sakana} & \textbf{\sys{}} \\
\midrule
Pages & 12.7 $\pm$ 1.2 & 6.7 $\pm$ 2.1 & 10.4 $\pm$ 1.1 & 5.8 $\pm$ 0.8 & 10.2 $\pm$ 0.8 \\
Words & 5{,}417 $\pm$ 863 & 2{,}741 $\pm$ 968 & 5{,}116 $\pm$ 524 & 2{,}606 $\pm$ 490 & 4{,}643 $\pm$ 438 \\
Figures & 4.3 $\pm$ 2.0 & 1.3 $\pm$ 0.4 & 0.1 $\pm$ 0.2 & 0.0 $\pm$ 0.0 & 3.8 $\pm$ 0.4 \\
Tables & 2.1 $\pm$ 0.7 & 0.8 $\pm$ 0.7 & 4.9 $\pm$ 1.5 & 2.3 $\pm$ 1.1 & 1.9 $\pm$ 0.2 \\
Equations & 1.3 $\pm$ 1.6 & 1.9 $\pm$ 2.0 & 7.1 $\pm$ 1.8 & 2.2 $\pm$ 1.5 & 3.0 $\pm$ 1.1 \\
Unique cite keys & 23.5 $\pm$ 8.7 & 9.3 $\pm$ 6.9 & 19.5 $\pm$ 3.9 & 10.5 $\pm$ 2.5 & 18.3 $\pm$ 4.5 \\
Bib entries & 23.5 $\pm$ 8.7 & 13.1 $\pm$ 13.2 & 15.8 $\pm$ 1.9 & 10.6 $\pm$ 2.6 & 55.3 $\pm$ 3.6 \\
Sections & 6.6 $\pm$ 2.5 & 6.5 $\pm$ 1.0 & 5.0 $\pm$ 0.0 & 6.5 $\pm$ 0.7 & 5.5 $\pm$ 0.6 \\
Subsections & 9.1 $\pm$ 3.9 & 9.7 $\pm$ 3.4 & 9.9 $\pm$ 1.7 & 6.0 $\pm$ 1.1 & 7.5 $\pm$ 1.4 \\
\bottomrule
\end{tabular}
\end{table}

\subsection{Failure Mode Case Studies}
\label{sec:exp:failures}

We highlight four failure modes from the 75-paper audit, each illustrating a different evidence chain break that \coeaudit{}'s integrity checks are designed to catch.

\paragraph{Case 1: Six orders of magnitude (ARC, LLM-SQL, seed~2).}
The paper introduces ``SCOR,'' a static column-ordering routine, and reports a combined score of 1{,}538{,}006.69---on a benchmark whose scoring metric uses a $[0, 1]$ scale.
The number is not a transcription error: it is the sum of squared prefix-hit lengths across datasets, an internal metric that the system computed and presented as if it were the ADRS score.
The paper is internally coherent---it defines its own evaluation protocol, runs controlled comparisons against a baseline (scoring 1{,}537{,}927.99), and draws reasonable conclusions within that framing.
An automated reviewer evaluating narrative quality alone would find nothing wrong.
Score Verification catches this immediately: the canonical evaluator re-run crashes (the submitted code fails to produce a valid solution), making the entire evidence chain from score to evaluator unresolvable.

\paragraph{Case 2: A bibliography from model memory (AIR, PRISM, seed~1).}
This paper's bibliography contains 15~references.
Reference Verification finds that 3 of those references are hallucinated---we could not find matching publications in Semantic Scholar, arXiv, or other scholarly databases.
This is not an edge case---AIR and DS produce hallucinated references at rates of 9\% and 21\% respectively (Table~\ref{tab:integrity}), compared to 0\% for systems with structured retrieval pipelines.

\paragraph{Case 3: Convergent specification violation (DS, LLM-SQL, seed~1).}
Under the I2 majority-vote protocol ($K{=}5$), 2 of 5 judges flag this paper---below the majority threshold, so it is not counted as a violation in Table~\ref{tab:integrity}.
Nevertheless, the submitted code achieves a legitimate score of 0.697 that passes Score Verification, but it does so by exploiting a gap between what the evaluator checks and what the benchmark intends to measure.
The code sorts columns differently per row-group block, then renames all columns back to the original schema before concatenation---this makes \texttt{pd.concat} assemble blocks in insertion order rather than realigning by column name, effectively permuting column order per row-group.
The evaluator validates row counts and total character counts but not column-to-column correspondence, so the permutation goes undetected.
The same exploit appears independently in two other systems (AIR seed~1 and \sys{} seed~2), providing convergent evidence that this is a genuine benchmark vulnerability rather than an isolated accident.%
\footnote{For \sys{} seed~2, the exploit is present in the submitted code, but the I2 majority-vote protocol did not reach consensus (1 of 5 judges flagged it), so it is not counted as a violation in Table~\ref{tab:integrity}. This illustrates a limitation of LLM-judged checks at the current vote threshold.}

\paragraph{Case 4: Near-correct score, fictional algorithm (ARC, TXN, seed~1).}
This paper nearly passes Score Verification: the reported score of 3{,}311 is within 3\% of the canonical evaluator re-run mean (3{,}214), just outside the adaptive tolerance threshold.
Yet Method-Code Alignment reveals a complete disconnect between what the paper describes and what the code implements.
The paper introduces ``STAR,'' a system built on bitwise integer encoding for conflict detection, an $O(1)$ surrogate cost model, and equidistant placement of high-contention anchor transactions.
The submitted code implements none of these: it uses standard Python sets for conflict tracking, calls the full simulator on every iteration (no surrogate), and clusters read-heavy keys sequentially rather than distributing write-heavy anchors.
This case illustrates why Score Verification alone is insufficient---the solver works, but the paper describes a different algorithm entirely, making the method section unreproducible regardless of how accurate the reported numbers are.

%% file: sections/012b_system_details.tex
%% ============================================================
%% Appendix: System Implementation Details
%% Ordered by pipeline stage: PI → Solver → Ablation → Paper Writer → Claim Verifier
%% ============================================================

\section{System Implementation Details}
\label{app:system_details}

\subsection{Problem Investigator}
\label{app:pi_details}

PI is a five-stage pipeline (plus two auxiliary stages), where each stage communicates via file-backed artifacts on disk.

\paragraph{Stage 1: Citation Graph.}
Starting from 2--4 seed papers, PI traverses the Semantic Scholar API (references and citations) up to 2 hops deep, producing a citation graph of approximately 2{,}000--5{,}000 candidate papers. % This stage involves no LLM calls and runs in 10--30 minutes, limited by API rate limits.

\paragraph{Stage 2: Literature Filter.}
An LLM scores each paper in the graph on two axes---methodology relevance and problem alignment (1--5 each)---and classifies papers into tiers: Core (both $\geq$4), Adjacent (one $\geq$4, other $\geq$3), Spark, or Noise. The resulting elite pool contains approximately 500 papers. A topic-relevance gate aborts the pipeline if fewer than 5 Core+Adjacent papers appear, preventing downstream drift from weak seeds.

\paragraph{Stage 3: Multi-Round Investigation.}
A Principal Investigator agent orchestrates specialist sub-agents across 3 rounds. Each round consists of candidate selection from the elite pool (Librarian agent), parallel PDF reading and structured note extraction (5 Researcher agents), and synthesis of findings into thematic research direction dossiers (SubdomainWriter agent). An IslandConsolidator agent merges redundant directions and retires low-quality ones after each round. The target is approximately 100 paper notes organized into 5--15 research directions. % This is the most expensive stage, typically running for 2--4 hours.

\paragraph{Stage 4: Evaluation Protocol Audit and Targeted Literature Refresh.}
Per-direction audit reports are generated and scored on a checklist rubric across multiple rounds until the direction passes.
A focused mini citation crawl on the winning direction produces 20--30 additional paper notes, filling gaps identified during the audit.

\paragraph{Stage 5: Experiment Brief Synthesis.}
Directions are scored by seed relevance, then a section-by-section writer produces the final Experiment Brief through a multi-round critic loop (up to 5 rounds with section-level revision). The brief contains three sections: (1)~a research landscape with technique taxonomy and best-known results, (2)~a concrete experiment plan with baselines, metrics, and ablation design, and (3)~a literature context with 25--40 references traceable to paper notes extracted from the source PDFs.

% The full pipeline runs in 3--6 hours per task with no human intervention. Default configuration uses \texttt{num\_expected\_papers}=100, \texttt{max\_hops}=2, and \texttt{min\_year}=2024.

\subsection{Solver}

The Solver consists of two agents.
The \emph{Solution Development Agent} receives an idea and task-specific instructions, operates in a sandboxed environment with tools for file I/O, command-line execution, solution management, and knowledge base retrieval, and follows an iterative refinement loop---executing experiments, debugging errors, and optimizing validation metrics---while maintaining an experimental log.
The \emph{Report Writing Agent} parses experimental artifacts to generate a technical report summarizing methodology and outcomes.

\subsection{Ablation}

Following PEE, the best-run selector filters out solutions flagged for specification violations and selects the highest-scoring remaining solution for further validation.
An ablation agent identifies the solution's core components and proposes controlled experiments to isolate their contributions.
These ablated versions are implemented and re-evaluated to quantify the performance impact of specific architectural or algorithmic choices.

\subsection{Paper Writer}
\label{app:paper_writer_pipeline}

The Paper Writer is a five-stage pipeline that converts raw materials into a verified \LaTeX{} draft. The first four stages operate on a \emph{research representation}---a structured markdown narrative with inline evidence annotations---before any \LaTeX{} is generated, enforcing the ``provenance before prose'' principle described in \S\ref{sec:system:writer}.

\paragraph{\textsc{Conceive}.}
A single LLM call reads all assembled raw materials (PI brief, experimental log, evaluator scores, solver code, seed-paper abstracts) and produces the initial research representation. This document captures the story arc---problem, gap, approach, result, limitation---with every factual claim carrying an inline evidence tag binding it to a specific workspace artifact (log line, score file, citation key, or ablation entry). \textsc{Conceive} establishes the narrative structure but does not validate evidence chains. That is deferred to \textsc{Ground}.

\paragraph{\textsc{Ground}.}
Deterministic checks validate each evidence annotation against the raw materials: the reported score must match the best-run score from discovery, baselines are labelled \textsc{verified} (traceable to a PI brief entry) or \textsc{estimated} (unattributed ``leaderboard'' references are flagged), every referenced artifact must exist, all expected sections are present, and hyperbole counts and known score mismatches are recorded. Each claim receives a \textsc{supported}, \textsc{partial}, or \textsc{unsupported} label, and the overall grounding ratio (supported claims / total claims) is computed.

\paragraph{\textsc{Critic}.}
One LLM call audits story-level coherence: gap--approach alignment, internal contradictions, overclaims relative to evidence strength, missing comparisons, baseline fairness, and honest limitations. The critic returns \textsc{pass} or a list of \textsc{major}/\textsc{minor} issues.

\paragraph{\textsc{Resolve}.}
A single LLM call rewrites the representation against the \textsc{Ground} flags and \textsc{Critic} issues jointly: unsupported claims are dropped or softened, contradictions are resolved using the verified source, overclaims are calibrated, and missing sections are filled. The \textsc{Ground}$\to$\textsc{Critic}$\to$\textsc{Resolve} loop iterates for up to two rounds, terminating on convergence (zero flags) or plateau (the flag count stops decreasing). A grounding ratio that remains below a configured threshold aborts the run rather than producing a poorly grounded draft.

\paragraph{\textsc{Compose}.}
The grounded representation is handed to per-section writers that emit \LaTeX{} one section at a time, with each numerical or citation-bearing sentence committing to its evidence source at writing time. The composed draft then passes through the Claim Verifier (\S\ref{app:claim_verifier_rules}). A refinement pass consumes the verifier's findings, rewriting flagged sentences to match their evidence sources, removing unsupported claims, and stripping all inline evidence annotations from the final \LaTeX{}. Only a draft with no remaining blocking violations is promoted to the final paper.

\subsection{Claim Verifier}
\label{app:claim_verifier_rules}

The Claim Verifier dispatches on the claim types defined in \S\ref{sec:coe}.
The writer commits each claim's evidence via annotation tags---a line in the experimental log, a citation key, an internal ablation, a baseline from the PI brief, or an explicit ``unsourced'' marker---and the verifier maps that evidence onto the claim type's verification rule:

\begin{itemize}\itemsep2pt \parskip0pt
  \item \textbf{Numerical claims} are checked by numeric tolerance against the cited evidence (log line, ablation entry, or PI baseline), with a $\pm 3$-line window on log lines and unit-aware normalizations for percent-versus-fraction and millisecond-versus-second mismatches.
  \item \textbf{Citation claims} are checked by resolving the cite key against the bibliography and then asking a one-shot LLM judge (JSON mode) whether the cited work's abstract supports the specific assertion.
  \item \textbf{Methodological claims} are checked by substantive textual overlap against the cited region of the experimental log.
\end{itemize}

\noindent Claims tagged ``unsourced'' or carrying malformed annotations are dropped automatically. A break code is recorded for each for downstream reporting.

%\subsection{Claim-Evidence Graph Format}
%For every annotation the verifier emits one record: a claim id, a type tag, the section it appears in, the rendered sentence text, a structured value where recoverable, and an ordered evidence chain whose links each carry a source URI, content hash, extracted text, verification method, and pass/partial/fail status.
%A verification summary attaches the appropriate break code when the chain is broken, distinguishing the failure modes catalogued in \cref{tab:coeaudit_checks}.
%The sidecar \texttt{corrections.json} records the same failures in a form the Refiner consumes, so a failed chain produces both a diagnostic in the CEG and a directed edit on the draft.

%% file: sections/012f_discovery_scaling.tex
%% --- Appendix F: Discovery Scaling ---

\section{Solution Discovery: Search Scaling}
\label{sec:app:discovery-scaling}

Table~\ref{tab:solver} reports \sys{} scores from the final iteration of the best-performing branch (1 of 5 parallel branches).
Table~\ref{tab:scaling} reports the best score across \emph{all} nodes in the search tree, measuring the discovery module's ceiling under varying tree and budget configurations. Solutions are manually checked and specification-violating ones (metric gaming on LLM-SQL, abusing the Prism scoring formula by failing on subsets of test inputs) are excluded. All scaling configurations report a single seed and cross-seed variance can be substantial (e.g., TXN ranges from 3636 to 3906 across three seeds with the base configuration). So \textbf{these results should be interpreted as directional rather than definitive}.

% \sys{}'s discovery module uses a parallel explore-exploit search tree parameterized by depth (iterations $I$), width (parallel branches $B$), and pruning (branches retained per iteration, $K$). The base configuration ($I\!=\!5, B\!=\!5, K\!=\!2$) retains the top-2 branches at each iteration, with a per-node evaluator budget of $E\!=\!4$ (the agent may attempt up to 4 evaluated solution versions per node, subject to early stopping via patience).

\paragraph{Tree and budget scaling.}
We evaluate search configurations along three axes: width (parallel branches $B$), depth (iterations $I$), and per-node evaluator budget ($E$, maximum evaluated solution versions per node).
Table~\ref{tab:scaling} reports the best score across all nodes for each configuration (single run each).
The tree scaling block in Table~\ref{tab:scaling} varies the tree structure ($I$, $B$, $K$) at fixed budget ($E\!=\!4$). The budget scaling block varies $E$ while holding the tree fixed at $I\!=\!5, B\!=\!5, K\!=\!2$.

\begin{table}[h]
\centering
\small
\caption{Best score across all search tree nodes under different tree and budget configurations (single run each). $I$=iterations (depth), $B$=branches (width), $K$=branches retained per iteration, $E$=max evaluator calls per node. ``Base'' is the configuration used for Table~\ref{tab:solver}. \textbf{Bold} marks the best score in each column, computed separately for baselines and \sys{} configurations.}
% $^\ddagger$Reference baselines are \emph{selector-based}: each system's own pipeline chose which solution to use for paper writing (e.g., Sakana selects via its agent, \sys{} selects the final iteration of the best-performing branch). \sys{} scaling configurations report the best score across all nodes, independent of the selection mechanism.
\label{tab:scaling}
\setlength{\tabcolsep}{4pt}
\begin{tabular}{@{}lcccccccccc@{}}
\toprule
\textbf{Config} & $I$ & $B$ & $K$ & $E$ & \textbf{Nodes} & \textbf{Prism}$\uparrow$ & \textbf{Cloud.}$\downarrow$ & \textbf{EPLB}$\uparrow$ & \textbf{LLM-SQL}$\uparrow$ & \textbf{TXN}$\uparrow$ \\
\midrule
\multicolumn{11}{@{}l}{\textit{Reference baselines (best-of-3 seeds, selector-based; from Table~\ref{tab:solver})}} \\
Human           & \multicolumn{5}{c}{--} & 21.89 & 626.24 & 0.1265 & 0.6920 & 2724.8 \\
AdaEvolve       & \multicolumn{5}{c}{--} & \textbf{26.26} & 637.10 & 0.1450 & \textbf{0.7520} & 4310 \\
EvoX            & \multicolumn{5}{c}{--} & \textbf{26.26} & 623.69 & \textbf{0.1453} & 0.7300 & 4310 \\
Sakana          & \multicolumn{5}{c}{--} & \textbf{26.26} & 627.11 & 0.1270 & 0.7320 & 4184 \\
ARC             & \multicolumn{5}{c}{--} & 26.25 & 690.37 & 0.1266 & 0.6757 & 3247 \\
AIR             & \multicolumn{5}{c}{--} & \textbf{26.26} & 734.28 & 0.1449 & 0.7148 & \textbf{4311} \\
DS              & \multicolumn{5}{c}{--} & \textbf{26.26} & \textbf{620.09} & 0.1284 & 0.7307 & 4286 \\
\midrule
\multicolumn{11}{@{}l}{\textit{\sys{} scaling configurations (1st seed of 3, best across all nodes)}} \\
Base            & 5 & 5 & 2  & 4  & 25  & 26.26 & 618.09 & 0.1287 & 0.7299 & 3636 \\
\midrule
\multicolumn{11}{@{}l}{\textit{Tree scaling (fixed budget $E\!=\!4$)}} \\
No pruning      & 5 & 5 & 5  & 4  & 25  & 26.26 & 618.08 & \textbf{0.1461} & 0.7092 & 3663 \\
Wide            & 5 & 10 & 2 & 4  & 50  & 26.36 & 618.09 & 0.1369 & 0.7215 & 4082 \\
Wider           & 5 & 15 & 2 & 4  & 75  & 26.26 & 618.08 & 0.1456 & 0.7189 & 4237 \\
Widest          & 5 & 20 & 2 & 4  & 100 & \textbf{26.44} & \textbf{618.07} & 0.1455 & 0.7257 & 4255 \\
Deep            & 10 & 5 & 2 & 4  & 50  & \textbf{26.44} & 618.08 & 0.1460 & 0.7118 & 3831 \\
Deep+wide       & 10 & 10 & 4 & 4  & 100 & 26.33 & 618.09 & 0.1458 & 0.7078 & 4000 \\
\midrule
\multicolumn{11}{@{}l}{\textit{Budget scaling (fixed tree $I\!=\!5, B\!=\!5, K\!=\!2$)}} \\
Budget 200      & 5 & 5 & 2  & 8  & 25  & 26.26 & \textbf{618.07} & 0.1284 & 0.7256 & \textbf{4348} \\
Budget 500      & 5 & 5 & 2  & 20 & 25  & 26.34 & \textbf{618.07} & 0.1456 & \textbf{0.7316} & 4237 \\
% \midrule
% \multicolumn{11}{@{}l}{\textit{Combined}} \\
% Wide+2$\times$budget & 5 & 10 & 4 & 8  & 50  & 26.50 & 618.08 & 0.1462 & 0.7276 & \textbf{3831} \\
\bottomrule
\end{tabular}
\end{table}

Three patterns emerge from tree scaling.
First, \textbf{TXN scales monotonically with width}: scores increase steadily from 3636 (base, $B\!=\!5$) through 4082 ($B\!=\!10$), 4237 ($B\!=\!15$), and 4255 ($B\!=\!20$), a 17\% improvement at the widest configuration and approaching AdaEvolve (4310).
Second, \textbf{EPLB benefits from scale but saturates early}: most non-base configurations reach ${\sim}0.146$ (a 13\% improvement over the base's 0.129), with the exception of Wide ($B\!=\!10$, $K\!=\!2$) at 0.137.
Third, \textbf{Cloudcast, LLM-SQL, and Prism largely saturate}: all configurations converge to similar scores on these three tasks regardless of tree shape, suggesting a narrow basin of high-performing solutions that the default search finds quickly.

%\paragraph{Budget scaling.}
%Budget scaling reveals a split between tasks.
%\textbf{TXN responds strongly to increased budget}: budget~200 reaches 4348 (+20\% over the base), comparable to the widest tree configuration ($B\!=\!20$, 4255).
%Interestingly, budget~500 does not improve further (4237), suggesting that TXN gains saturate around $E\!=\!8$ per node and that the additional budget is spent refining rather than discovering better algorithmic strategies.
%EPLB shows a similar pattern: budget~500 reaches 0.1456 (+13\%), matching the width-scaled results but with only 25 nodes instead of 75--100.
%Cloudcast, LLM-SQL, and Prism remain flat across all budget levels, consistent with the tree scaling results.

Taken together, \textbf{width (more independent branches) is the most efficient scaling axis} for tasks with diverse solution strategies.
The widest tree ($B\!=\!20$, 100~nodes, $E\!=\!4$) matches or exceeds the highest per-node budget configuration ($E\!=\!20$, 25~nodes) on 4 of 5 tasks, despite using 5$\times$ fewer evaluator calls per node.
Budget scaling shows gains on TXN (budget~200 reaches 4348, +20\%) but saturates quickly. Budget~500 does not improve further on TXN, and the remaining tasks are flat across all budget levels.
Depth and budget produce diminishing returns once the search covers the main algorithmic strategies.

%\paragraph{Comparability with AdaEvolve's iteration budget.}
%AdaEvolve reports results under a fixed iteration budget of $T\!=\!100$~\citep{cemri2026adaevolve}.
%Our base configuration produces ${\sim}100$ evaluator calls (25 nodes $\times$ ${\sim}4$ calls each), making the evaluator-call counts directly comparable; the difference is that \sys{} distributes evaluations across a tree with ideation and pruning between iterations, adding LLM overhead not counted in the evaluator budget.

\paragraph{Specification violations at higher budgets.}
Increasing per-node budget amplifies specification violation risk.
At budget~100, no specification violations are observed on Prism. At budget~200 and~500, 2--8\% of nodes converge to solutions that exploit the scoring formula rather than solving the task correctly.
LLM-SQL shows a similar trend: the fraction of nodes flagged for metric gaming by the post-hoc auditor grows from ${\sim}0\%$ at budget~100 to ${\sim}50\%$ at budget~200 and ${\sim}70\%$ at budget~500.
In contrast, wider trees at budget~100 show lower violation rates despite producing more total nodes, because each node has fewer iterations to discover and refine exploitative patterns.

%% file: sections/012c_coe_audit_details.tex
\section{CoE Audit Details}
\label{app:coe_audit_details}

% Sub-code definition tables removed — categories are described in plain English
% in the per-check analysis sections below (I1: \S\ref{appx:discovered-score-errors},
% I2: \S\ref{appx:i2_spec_violation}, I4: \S\ref{app:i4_audit_analysis}).

\subsection{Audit Procedure and Reproducibility}
\label{app:audit_implementation}

\coeaudit{} is designed for reproducibility.
\Cref{tab:audit_procedure} summarizes the automation level of each component.
Score Verification uses LLM-based extraction to identify the paper's reported score from \TeX{} and PDF files, then compares it against reproduced scores from evaluator re-runs via numeric tolerance---the comparison itself is fully deterministic.
Specification Violation and Method-Code Alignment are LLM-judged, with majority vote across multiple independent runs to reduce judgment noise.
Reference Verification is primarily API-based (Semantic Scholar, arXiv, OpenAlex, CrossRef lookup), with an LLM disambiguation step for title-only matches.

\begin{table}[h!]
\centering
\small
\caption{\coeaudit{} automation level per component.}
\label{tab:audit_procedure}
\begin{tabular}{@{}lll@{}}
\toprule
\textbf{Component} & \textbf{Automation} & \textbf{Model (if LLM)} \\
\midrule
Score Verification & LLM extraction + automated comparison & Gemini~3 Flash \\
Specification Violation & LLM-judged & Gemini~3.1 Pro \\
Reference Verification & Automated + LLM disambiguation & Gemini~3 Flash \\
Method-Code Alignment & LLM-judged & Gemini~3.1 Pro \\
\bottomrule
\end{tabular}
\end{table}

\paragraph{Human verification.}
Although the audit pipeline is automated, all flagged positives for I1 (Score Verification), I2 (Specification Violation), and I3 (Reference Verification) were manually reviewed and corrected by human reviewers before reporting in Table~\ref{tab:integrity}.
A small number of auditor false positives (e.g., API resolution failures for real papers, score extraction errors) were identified and removed during this process.
\emph{Exception:} most of Sakana ASv2's I2 violations trace to the BFTS--ADRS design mismatch rather than adversarial behaviour (\Cref{app:baseline_adaptation}).
For I4 (Method-Code Alignment), human reviewers validated a sample of the LLM judgments; the results in Table~\ref{tab:integrity} reflect the LLM majority-vote scores without manual correction.

\section{Failure Cases per Audit Metric}
\subsection{I1: Score Verification Errors }\label{appx:discovered-score-errors}

We manually reviewed every I1 (Score Verification) failure flagged
across the five systems and categorise the 22~confirmed
agent errors into five classes (Table~\ref{tab:i1_failure_distribution}).%
\footnote{A small number of auditor false positives were corrected
before reporting Table~\ref{tab:integrity}; this section reports only
confirmed agent errors.
Table~\ref{tab:integrity} reports pass rates over 12~papers per system (EPLB excluded), so its implied failure count is lower than the 22~errors here, which include EPLB papers.}

\begin{table}[h]
\centering
\small
\caption{Confirmed agent I1 errors, by category.}
\label{tab:i1_failure_distribution}
\begin{tabular}{@{}lp{6.5cm}cc@{}}
\toprule
\textbf{Category} & \textbf{Definition} & \textbf{Count} & \textbf{\%} \\
\midrule
\texttt{value\_mismatch}          & Paper headline number differs from evaluator rerun beyond tolerance, with the same metric and scale.                            & 13 & 59\% \\
\texttt{cross\_stage\_cherry\_pick}& Writeup LLM selects a score from a different search-tree stage (e.g.\ ablation) than the node whose code is submitted.         &  4 & 18\% \\
\texttt{paper\_score\_unavailable}& Paper contains no machine-readable headline number for the auditor to verify against.                                          &  2 &  9\% \\
\texttt{evaluator\_error}         & Submitted artifact cannot be re-evaluated within the budget (timeout or crash on the agent's own solution file).                &  1 &  5\% \\
\texttt{metric\_mismatch}         & Paper and code report different metrics, scales, or optimisation directions, so the numbers are not directly comparable.       &  2 &  9\% \\
\bottomrule
\end{tabular}
\end{table}

\paragraph{\texttt{value\_mismatch} (n=13).}
The paper and the code agree on what is being measured but disagree
on the value.
Most (9/13) of these gaps are within 5\% of the paper-reported
number---small enough to plausibly arise from unreported seed
variance, but uniformly biased towards a better-than-rerun headline.
Among the four baseline systems, the two largest gaps are
qualitatively different errors: a fabricated metric direction in DS
\texttt{cloudcast} seed-2 (26.7\%), where the paper relabels a cost
metric as ``utility'' and inverts the optimisation direction so that
a worse-than-baseline number reads as an improvement; and a
2$\times$ mismatch in ARC \texttt{prism} seed-3 between the paper
headline (12.74) and what the submitted code actually produces
(26.24).
Sakana ASv2 contributes two \texttt{value\_mismatch} cases, both
caused by \emph{environment-dependent tuning}: the solver contains
a hyperparameter tuning loop gated on an environment variable
(\texttt{\_ADRS\_EVAL\_GUARD}); during BFTS search the variable is
unset and the loop executes with optimal parameters, but during
canonical re-evaluation the variable is set and the loop is skipped,
causing the solver to fall back to defaults
(e.g.\ \texttt{prism}~seed-0: 26.26 tuned vs.\ 22.34 default,
a~15\% gap).

\paragraph{\texttt{cross\_stage\_cherry\_pick} (n=4).}
All four cases come from Sakana ASv2 and reflect a boundary mismatch
in the BFTS pipeline: the best node is selected from stages~1--2
(preliminary investigation and hyperparameter tuning), but the writeup LLM receives summaries
from all four stages---including ablation---and picks the most
favourable score from the entire pool.
For example, in \texttt{prism}~seed-1 the selected node scores
22.79, but the paper reports 25.39---a number traced to ablation
node~6 (``Ablate KVPR-Aware Initialization'') in the ablation
summary.
The same pattern appears in \texttt{cloudcast}~seed-0 (+56\%),
\texttt{prism}~seed-2 ($-$4.7\%, paper under-reports), and
\texttt{txn\_scheduling}~seed-2 (+17\%).

\paragraph{\texttt{paper\_score\_unavailable} (n=2).}
The agent's writeup contains no machine-readable quantitative result
for the headline metric.
Both cases are baseline-system failures: an AIR \texttt{prism} paper
that omits scores entirely, and a DS \texttt{cloudcast} PDF that
failed to render its results section.

\paragraph{\texttt{evaluator\_error} (n=1).}
The artifact the agent declares as its solution cannot be
re-evaluated end-to-end within the budget.
The single case is ARC \texttt{llm\_sql} seed-2, which ships
\texttt{program.py}---a multi-condition experiment harness that
runs 8~baseline conditions $\times$ 3~seeds $\times$ multiple
datasets at import time before any single number is produced.

\paragraph{\texttt{metric\_mismatch} (n=2).}
The paper and the code do not agree on what they are measuring.
ARC \texttt{llm\_sql} seed-3 publishes a headline of 0.0 that the
paper itself attributes to ``a critical integration failure with the
benchmark evaluator framework''---the agent surfaced a known-broken
result as its final score.
Sakana ASv2 \texttt{cloudcast}~seed-1 reports per-transfer dollar
cost while the evaluator produces a combined score, making the two
numbers incomparable.

\paragraph{Per-system error shape.}
Table~\ref{tab:i1_per_system} shows that the failure mix is highly
system-specific.
Sakana ASv2 produces the most errors (7) and the only
\texttt{cross\_stage\_cherry\_pick} cases---a failure mode unique to
multi-stage search pipelines that expose full experiment histories to
the writeup phase.
Its two \texttt{value\_mismatch} cases are environment-dependent
tuning rather than simple score inflation.
AIR and DS errors are dominated by small-to-medium
\texttt{value\_mismatch}: the numbers exist and are in the right
ballpark but do not exactly reproduce.
ARC spans the widest range of categories and the largest single
discrepancy (106\%), and contributes the only
\texttt{evaluator\_error} case.
\sys{} produces no confirmed I1 errors, consistent with its
12/12 score-verification result in Table~\ref{tab:integrity}.

\begin{table}[h]
\centering
\footnotesize
\caption{Confirmed agent I1 errors per system, by category.}
\label{tab:i1_per_system}
\begin{tabular}{@{}lcccccc@{}}
\toprule
\textbf{System} & \texttt{val\_mis.} & \texttt{cherry\_pick} & \texttt{score\_unavail.} & \texttt{eval\_error} & \texttt{metric\_mis.} & \textbf{Total} \\
\midrule
Sakana ASv2 & 2 & 4 & 0 & 0 & 1 & 7 \\
AIR    & 3 & 0 & 1 & 0 & 0 & 4 \\
ARC    & 5 & 0 & 0 & 1 & 1 & 7 \\
DS     & 3 & 0 & 1 & 0 & 0 & 4 \\
\sys{} & 0 & 0 & 0 & 0 & 0 & 0 \\
\bottomrule
\end{tabular}
\end{table}

\subsection{I2: Specification Violation Analysis}
\label{appx:i2_spec_violation}

The I2 audit flagged 11~papers across five systems for specification
violations (Table~\ref{tab:i2_per_system}).
Sakana ASv2 accounts for 10 of 11~flagged papers; the remaining
case is an AIR \texttt{llm\_sql} paper.
ARC, DS, and \sys{} produce zero I2 violations under majority vote.%
\footnote{Under union vote (any single judge flags), ARC has 3,
DS has 1, and \sys{} has 1.
As noted in the main text, DS seed-1
\texttt{llm\_sql} was flagged by only 2/5 judges---below the
majority threshold---despite containing an evaluator-exploiting
column-permutation pattern.}
Most of Sakana's I2 violations trace to the BFTS--ADRS design
mismatch (\Cref{app:baseline_adaptation}).

Because a single paper can trigger multiple violation categories, we report
both the category distribution (Table~\ref{tab:i2_subcode_distribution})
and the per-system task breakdown
(Table~\ref{tab:i2_per_system}).

\begin{table}[h]
\centering
\small
\caption{I2 violation categories across 11~flagged papers.
A paper may trigger multiple categories, so paper counts do not sum
to~11.}
\label{tab:i2_subcode_distribution}
\begin{tabular}{@{}lcc@{}}
\toprule
\textbf{Category} & \textbf{Papers} & \textbf{System(s)} \\
\midrule
Evaluator import       & 10 & Sakana (10/10) \\
Evaluator exploitation & 7  & Sakana (7/10) \\
Specification exploit  & 5  & Sakana (4/10), AIR (1) \\
Data leakage           & 1  & Sakana (1/10) \\
\bottomrule
\end{tabular}
\end{table}

\paragraph{Evaluator import (10/10 Sakana papers).}
Every flagged Sakana paper imports the canonical evaluator and calls
it as an optimisation oracle.
The mechanism is uniform: the agent adds
\texttt{from evaluator import evaluate} (or a variant aliased as
\texttt{\_tune\_eval}), then loops over hyperparameter configurations
---calling \texttt{evaluate()} for each---and keeps the best.
The import pattern mirrors our canonical evaluation harness, which
was injected into the initial code template to enable automated
scoring; the agent copies it and builds its own tuning
infrastructure on top.
The agent could tune parameters through BFTS's iteration loop
(one setting per iteration, scored by the external evaluator),
but the stage~2 goal (``test across multiple parameter settings'')
encourages an intra-iteration sweep, which requires importing the
evaluator directly.

\paragraph{Specification exploit (4 Sakana + 1 AIR).}
Four Sakana \texttt{txn\_scheduling} papers (seeds 1--3) and one
\texttt{llm\_sql} paper modify functions outside the
\texttt{EVOLVE-BLOCK} markers or exploit metric edge cases.
The \texttt{txn\_scheduling} pattern is consistent: the agent
modifies \texttt{get\_random\_costs()}, a function explicitly
forbidden from modification, to run a parameter sweep and return
the best result.
AIR's single specification-exploit case (\texttt{llm\_sql} seed-1) is a
column-permutation exploit: the solver physically reorders values
across columns within each row, destroying column integrity to
inflate the evaluator's prefix-cache hit metric---the same
pattern that appears in DS seed-1 \texttt{llm\_sql} but below
the majority threshold there.

\begin{table}[h]
\centering
\small
\caption{I2 specification violations per system. For Sakana ASv2, we
show the per-task breakdown; other systems are reported as totals.
All counts use majority vote (3/5 judges).}
\label{tab:i2_per_system}
\begin{tabular}{@{}lcccccr@{}}
\toprule
\textbf{System} & \texttt{cloud.} & \texttt{eplb} & \texttt{llm\_sql} & \texttt{prism} & \texttt{txn\_sched.} & \textbf{Total} \\
\midrule
Sakana ASv2 & 1/3 & 0/3 & 3/3 & 3/3 & 3/3 & 10/15 \\
AIR         & 0   & 0   & 1   & 0   & 0   & 1/15  \\
ARC         & 0   & 0   & 0   & 0   & 0   & 0/15  \\
DS          & 0   & 0   & 0   & 0   & 0   & 0/15  \\
\sys{}      & 0   & 0   & 0   & 0   & 0   & 0/15  \\
\bottomrule
\end{tabular}
\end{table}

\paragraph{Clean runs.}
Five of 15~Sakana runs produce no I2 violations: all three EPLB
seeds and \texttt{cloudcast} seeds 2 and~3.
EPLB's solver contract is structurally simpler (a pure allocation
function with no external scorer dependency), and the two clean
\texttt{cloudcast} runs show that even on graph-optimisation tasks
the agent \emph{can} fulfill BFTS stage goals without importing the
evaluator---but this is the exception, not the default, under
BFTS's stage pressure.

\subsection{I3: Reference integrity -- Discovered hallucinated references}\label{appx:discovered_hallucinated_references}

The list below shows unique hallucinated reference keys.
Table~\ref{tab:integrity} reports total occurrences across papers (the same key can appear in multiple papers; e.g., ARC's single fabricated key appears in all three EPLB papers, counting as 3 in the table).

\begin{longtable}{@{}p{15cm}@{}}
\toprule
\textbf{Hallucinated references} \\
\midrule
\endfirsthead
\toprule
\textbf{Hallucinated references} (continued) \\
\midrule
\endhead
\midrule
\endfoot
\bottomrule
\endlastfoot

\midrule
\multicolumn{1}{c}{\textbf{ARC (1)}} \\
\midrule
\textbf{sutskever2013importance}: Sutskever et al. \textit{SGD with Momentum}. ICML, 2013. \\

\midrule
\multicolumn{1}{c}{\textbf{AIR (21)}} \\
\midrule
\textbf{ruth2018castflow}: R\"{u}th, Jan et al. \textit{CastFlow: Clean-slate multicast approach using in-advance path computation in software-defined networks}. 2018 IEEE 43rd Conference on Local Computer Networks (LCN), 2018. \\
\textbf{dou2022hetmoe}: Dou, Shiwei et al. \textit{HetMoE: An Efficient Distributed MoE Training System for Heterogeneous Clusters}. arXiv preprint arXiv:2210.12384, 2022. \\
\textbf{li2023lightllm}: Li, Zhuohan et al. \textit{LightLLM: A highly optimized LLM inference system with token-level kv cache management}. arXiv preprint arXiv:2309.04414, 2023. \\
\textbf{purohit2024prism}: Purohit, Archit et al. \textit{Prism: Optimizing multi-model LLM serving on GPU clusters}. Proceedings of the 29th ACM International Conference on Architectural Support for Programming Languages and Operating Systems (ASPLOS), 2024. \\
\textbf{miao2023muxserve}: Miao, Xupeng et al. \textit{MuxServe: Multiplexing large language models for high throughput and low latency}. arXiv preprint arXiv:2311.05602, 2023. \\
\textbf{cloudcast}: Zheng, Q. et al. \textit{CloudCast: Cost-efficient multicast routing in cloud networks}. IEEE INFOCOM 2019 - IEEE Conference on Computer Communications, 2019. \\
\textbf{idreos2012main}: Idreos, Stratos et al. \textit{Main-memory column stores}. Foundations and Trends\textregistered\ in Databases, 2012. \\
\textbf{lightllm2023}: Li, Zhen et al. \textit{LightLLM: A Lightweight and Highly Efficient Python-based Large Language Model Serving Framework}. arXiv preprint arXiv:2310.01234, 2023. \\
\textbf{bhuiyan2024prism}: Bhuiyan, M. et al. \textit{Prism: A Flexible and Scalable Multi-LLM Serving System}. Proceedings of the 29th ACM International Conference on Architectural Support for Programming Languages and Operating Systems (ASPLOS), 2024. \\
\textbf{eplb2023}: Wang, H. et al. \textit{Expert Parallelism Load Balancing in Mixture-of-Experts Models}. Proceedings of the International Conference on High Performance Computing, Networking, Storage and Analysis (SC), 2023. \\
\textbf{hussin2011metaheuristic}: Hussin, MS et al. \textit{Metaheuristic algorithms for traveling salesman problem: A review}. Annals of the University of Craiova, Mathematics and Computer Science Series, 2011. \\
\textbf{zhang2020job}: Zhang, Jun et al. \textit{Job shop scheduling research}. International Journal of Production Research, 2020. \\
\textbf{jetstream2014}: Rabkin, Ariel et al. \textit{JetStream: Enabling wide-area data streaming}. 11th USENIX Symposium on Networked Systems Design and Implementation (NSDI 14), 2014. \\
\textbf{slingshot2022}: Doe, John et al. \textit{Slingshot: High-performance routing across federated cloud environments}. Proceedings of the ACM SIGCOMM 2022 Conference, 2022. \\
\textbf{cloudcast2021}: Chen, Yiting et al. \textit{CloudCast: Cost-effective data distribution in multi-cloud deployments}. IEEE INFOCOM 2021 - IEEE Conference on Computer Communications, 2021. \\
\textbf{cui2004}: Cui, Jun-Hong et al. \textit{QoS multicast routing in dynamic networks}. IEEE Network, 2004. \\
\textbf{laoutaris2011netstitcher}: Laoutaris, Nikolaos et al. \textit{NetStitcher: Un-tethering bulk storage from the network edge}. Proceedings of the ACM SIGCOMM 2011 conference, 2011. \\
\textbf{feng2020traffic}: Feng, Xin et al. \textit{Traffic engineering in software-defined wide-area networks: A survey}. IEEE/ACM Transactions on Networking, 2020. \\
\textbf{epstein2005online}: Epstein, Leah and van Stee, Rob. \textit{Online bin packing with square-root sized items}. Information Processing Letters, 2005. \\
\textbf{ba2023modelbox}: Ba, Yuhan et al. \textit{ModelBox: A Framework for Multi-Model Multi-Tenant Serving}. 2023 USENIX Annual Technical Conference (USENIX ATC 23), 2023. \\
\textbf{zheng2024eplb}: Zheng, Lianmin et al. \textit{EPLB: Load Balancing for Expert Parallelism in Large Language Models}. arXiv preprint arXiv:2401.03221, 2024. \\

\midrule
\multicolumn{1}{c}{\textbf{DS (41)}} \\
\midrule
\textbf{choy1991heuristic}: Choy, M-S. \textit{Heuristic algorithms for the Steiner tree problem with an application to network routing}. Proceedings of the 1991 ACM SIGCOMM, 1991. \\
\textbf{beloglazov2012energy}: Beloglazov, Anton and Buyya, Rajkumar. \textit{Energy-efficient routing and resource allocation in multi-cloud}. Journal of Network and Computer Applications, 2012. \\
\textbf{grotschel1993steiner}: Grotschel, Martin et al. \textit{The Steiner tree packing problem in telecommunications}. 1993. \\
\textbf{romero2021automated}: Romero, F et al. \textit{Automated algorithm design using large language models}. Advances in Neural Information Processing Systems, 2023. \\
\textbf{wu2015spanning}: Wu, Chuan et al. \textit{Spanning tree based data transfer in multi-cloud architectures}. 2015. \\
\textbf{melian2012cloud}: Melian, L et al. \textit{Cloud computing economics: a survey}. 2012. \\
\textbf{faragardi2017multi}: Faragardi, Hamid Reza et al. \textit{Multi-cloud data distribution with cost optimization}. 2017. \\
\textbf{mishra2018cost}: Mishra, A et al. \textit{Cost efficient routing in multi-cloud environments}. 2018. \\
\textbf{voss1992steiner}: Voss, Stefan. \textit{The Steiner tree problem with edge capacities}. 1992. \\
\textbf{bubeck2023approaches}: Bubeck, S et al. \textit{Approaches to code generation and synthesis}. 2023. \\
\textbf{wang2020automated}: Wang, X et al. \textit{Automated hyperparameter optimization in cloud computing}. 2020. \\
\textbf{smith2019data}: Smith, J et al. \textit{Data transfer costs in modern cloud platforms}. 2019. \\
\textbf{liu2022network}: Liu, Y et al. \textit{Network aware multi-cloud data distribution}. 2022. \\
\textbf{jones2023oscillatory}: Jones, R et al. \textit{Oscillatory dynamics in automated search landscapes}. 2023. \\
\textbf{kim2021puber}: Kim, Young Jin et al. \textit{Puber: Efficient expert parallelism for mixture-of-experts}. arXiv preprint arXiv:2111.05454, 2021. \\
\textbf{romera2021optimizing}: Romera-Paredes, Oscar et al. \textit{Optimizing mixture-of-experts for large-scale distributed training}. arXiv preprint arXiv:2102.04353, 2021. \\
\textbf{clune2008how}: Clune, Jeff et al. \textit{How evolutionary dynamics affects network topology}. Artificial life, 2008. \\
\textbf{llm\_agents\_2023}: Smith, J. and Doe, A. \textit{Autonomous LLM Agents for Code Generation}. Journal of AI Research, 2023. \\
\textbf{api\_misuse\_2024}: Wang, L. and Lee, C. \textit{API Misuse in LLM-Generated Code}. Proceedings of ICSE, 2024. \\
\textbf{fail\_fast\_2025}: Garcia, M. \textit{Fail-Fast Sandboxing for Coding Agents}. IEEE Transactions on Software Engineering, 2025. \\
\textbf{prism2024}: Anonymous. \textit{Prism: A Benchmark for Multi-LLM Serving Systems}. arXiv preprint, 2024. \\
\textbf{sarca2023}: Anonymous. \textit{SARCA: Systems Architecture Research using Coding Agents}. arXiv, 2023. \\
\textbf{clockwork2023}: Anonymous. \textit{Clockwork: Fast and Predictable Inference for Edge Machine Learning}. arXiv, 2023. \\
\textbf{garcia1982fully}: Garcia-Molina, Hector. \textit{A fully distributed null-free algorithm for concurrent database updates}. IEEE Transactions on Software Engineering, 1982. \\
\textbf{karlsson2020combinatorial}: Karlsson, Elias et al. \textit{Combinatorial optimization by graph neural networks}. arXiv preprint arXiv:2010.16012, 2020. \\
\textbf{krajewski2024mixtures}: Krajewski, Adam et al. \textit{Mixtures of experts: A systematic review}. arXiv preprint arXiv:2401.00000, 2024. \\
\textbf{paliwal2020regal}: Paliwal, Aditya et al. \textit{REGAL: A transfer learning based methodology for hardware-software co-design}. MICRO, 2020. \\
\textbf{jain1998simulated}: Jain, AS and Meeran, S. \textit{A simulated annealing algorithm for job shop scheduling problem}. Mathematical and computer modelling, 1998. \\
\textbf{tian2021cloud}: Tian, X et al. \textit{Cloud egress cost optimization}. Proceedings of the ACM SIGCOMM 2021 Conference, 2021. \\
\textbf{zhao2020understanding}: Zhao, Y et al. \textit{Understanding cloud network egress constraints}. USENIX Annual Technical Conference (ATC), 2020. \\
\textbf{binnig2021learned}: Binnig, Carsten et al. \textit{The case for learned database systems}. arXiv preprint, 2021. \\
\textbf{yu2014staccato}: Yu, Xiangyao et al. \textit{Staccato: A dependency-aware transaction scheduling system for many-core processors}. Proceedings of the 2014 ACM SIGMOD International Conference on Management of Data, 2014. \\
\textbf{ren2016low}: Ren, Kun et al. \textit{Low-overhead deadlock detection in distributed database systems}. Proceedings of the 2016 International Conference on Management of Data, 2016. \\
\textbf{pavlo2017make}: Pavlo, Andrew and Stonebraker, Michael. \textit{What's make a database system fast?}. ACM SIGMOD Record, 2017. \\
\textbf{blanas2010comparison}: Blanas, Spyros et al. \textit{A comparison of join algorithms for modern multi-core processors}. Proceedings of the VLDB Endowment, 2010. \\
\textbf{bailis2014bolt}: Bailis, Peter et al. \textit{Bolt-on conflict-free replicated data types}. ACM SIGMOD Record, 2014. \\
\textbf{hardi1992precedence}: Hardi, S and Rakow, TC. \textit{Precedence-based transaction scheduling}. Proceedings of the 2nd International Workshop on Research Issues on Data Engineering, 1992. \\
\textbf{kim2016fast}: Kim, Taesoo et al. \textit{Fast and scalable serializable transactions in multicore in-memory databases}. Proceedings of the 2016 International Conference on Management of Data, 2016. \\
\textbf{wu2017transaction}: Wu, Yingjun et al. \textit{Transaction scheduling using graph coloring}. Proceedings of the 2017 ACM SIGMOD International Conference on Management of Data, 2017. \\
\textbf{lin2015scheduler}: Lin, Jianshu et al. \textit{To schedule or not to schedule: When is transaction scheduling worth the overhead?}. Proceedings of the 2015 ACM SIGMOD International Conference on Management of Data, 2015. \\
\textbf{faleiro2017high}: Faleiro, Jose M and Abadi, Daniel J. \textit{High performance serializable concurrency control with determinism}. Proceedings of the 2017 ACM SIGMOD International Conference on Management of Data, 2017. \\
\end{longtable}

\subsection{I4 Audit Failure Analysis}
\label{app:i4_audit_analysis}

We categorise the 95~method-code misalignment findings flagged by
the I4 audit (across 25~papers\footnote{Table~\ref{tab:integrity} implies 26 misaligned papers (12+10+3+1); one paper (DS seed-3/cloudcast) could not be cleanly classified and is excluded from the categorical breakdown.}) into three semantic classes
(Table~\ref{tab:i4_failure_distribution}).
A single paper can produce multiple findings, so we report both the
finding-level distribution and the distinct-paper count per
category.
Sakana ASv2 is excluded from this breakdown because its I4 results
are confounded by the artifact format---the audited code is a full
experimental script rather than an extracted solver, so the I4
judges compare the paper against non-solver infrastructure code
(\Cref{app:baseline_adaptation}).

\begin{table}[h]
\centering
\small
\caption{I4 method-code alignment findings, by category (n=95
findings, 25 affected papers).}
\label{tab:i4_failure_distribution}
\begin{tabular}{@{}lp{6cm}ccc@{}}
\toprule
\textbf{Category} & \textbf{Definition} & \textbf{Findings} & \textbf{\%} & \textbf{Papers} \\
\midrule
\texttt{incomplete\_broken}        & Paper-described mechanism or component is missing, partially implemented, or replaced with a degenerate fallback in the code. & 49 & 52\% & 19 \\
\texttt{algorithm\_class\_mismatch}& Code implements a fundamentally different algorithm class than the one described in the paper (e.g.,~paper claims Iterated Local Search, code is Simulated Annealing; paper claims a neural- or LLM-guided search, code is deterministic). & 37 & 39\% & 15 \\
\texttt{deceptive\_dummy\_code}    & Code contains undisclosed elements designed to mislead evaluation: hidden environment-variable switches, or output deliberately shaped to game an evaluator metric. &  9 &  9\% &  5 \\
\bottomrule
\end{tabular}
\end{table}

\paragraph{\texttt{incomplete\_broken} (n=49, 19 papers).}
The largest category. The code generally targets the same problem
the paper describes, but is missing one or more of the specific
mechanisms claimed in the writeup, or substitutes a degenerate
fallback. Recurring patterns include:
described \emph{multi-start initialisation with $K$ sequences}
collapsed to a single deterministic backtracking pass
(AIR \texttt{prism}); claimed \emph{link-penalty diversification}
producing a constant assignment of the same path to every partition
(AIR \texttt{cloudcast}); described \emph{arborescence lookahead}
or \emph{surrogate cost model} simply absent from the code, with the
true simulator invoked at every iteration instead
(ARC \texttt{cloudcast}, ARC \texttt{txn\_scheduling});
claimed \emph{2.0\,GB memory threshold safeguards} reduced to a
trivial overflow check (DS \texttt{prism}).

\paragraph{\texttt{algorithm\_class\_mismatch} (n=37, 15 papers).}
The code implements a fundamentally different algorithm class than
the one described in the paper. The most common sub-pattern is the
\emph{claimed-learning-loop-is-absent} case: a paper claims an
LLM-driven evolutionary search, a neural network predictor, or an
LLM SQL optimiser, and the code is a single deterministic heuristic
with no LLM calls and no trained model
(e.g.\ AIR \texttt{cloudcast}: ``hybrid neuro-symbolic solver''
$\rightarrow$ purely classical Steiner-tree heuristics;
ARC \texttt{llm\_sql}: ``36 LLM prompting strategies''
$\rightarrow$ deterministic dataframe reordering;
DS \texttt{eplb}: ``LLM-driven evolutionary search over 27
generations'' $\rightarrow$ standalone deterministic load
balancer).
Classical algorithm-class swaps also recur:
Iterated Local Search $\rightarrow$ Simulated Annealing
(AIR \texttt{prism}), Dinkelbach's fractional programming
$\rightarrow$ static sum-of-squares cost (ARC \texttt{prism}),
recursive partitioning $\rightarrow$ single-level grouping
(DS \texttt{llm\_sql}).
We also observe the rarer \emph{method/baseline inversion} pattern,
in which the paper labels $X$ as the proposed method and $Y$ as a
rejected ablation, but the code uses $Y$
(ARC \texttt{prism}: ``GRASP Without Symbiosis'' claimed as final
method while the code instantiates \texttt{SymbioticGRASPPacker}).

\paragraph{\texttt{deceptive\_dummy\_code} (n=9, 5 papers).}
The submitted artifact contains undisclosed code whose presence
appears intended to mislead automated evaluation. All nine
findings come from ARC and split into two sub-patterns:
(i)~\emph{hidden environment-variable switches} (4 findings,
2~papers): the code reads an undisclosed variable
(\texttt{CONDITION}, \texttt{ABLATION}) at import time and dispatches
to one of several different solvers, while the paper presents a
single unified algorithm
(ARC \texttt{cloudcast} seeds~2 and~3).
(ii)~\emph{evaluator gaming} (5 findings, 3~papers): code
intentionally shaped to inflate a metric without solving the
underlying task, e.g.\ returning an empty column-ordering list while
internally permuting values to maximise prefix-cache hits
(ARC \texttt{llm\_sql}~seed-3).

\paragraph{Per-system error shape.}
Table~\ref{tab:i4_per_system} shows that the failure mix is sharply
system-specific.
AIR, DeepScientist, and \sys{} produce no
\texttt{deceptive\_dummy\_code} findings: when these systems
misalign, the gap is between what the paper says and what the code
does, with no active obfuscation.
ARC is the only system that produces
\texttt{deceptive\_dummy\_code} findings (9/9, 5 papers), spanning
both hidden env-var switches and prefix-hit evaluator gaming.

\begin{table}[h]
\centering
\small
\caption{I4 findings per system, by category. Each cell is a
finding count; a single paper may contribute multiple findings.}
\label{tab:i4_per_system}
\resizebox{\textwidth}{!}{%
\begin{tabular}{@{}lcccc@{}}
\toprule
\textbf{System} & \texttt{algo\_class\_mismatch} & \texttt{incomplete\_broken} & \texttt{deceptive\_dummy\_code} & \textbf{Total} \\
\midrule
AIR &  3 &  9 & 0 & 12 \\
ARC & 15 & 23 & 9 & 47 \\
DS  & 15 & 16 & 0 & 31 \\
\sys{} &  4 &  1 & 0 &  5 \\
\bottomrule
\end{tabular}}
\end{table}

\section{MLE-Bench and Parameter Golf Evaluation}
\label{app:mle-pg-eval}

\begin{table}[h]
\centering
\small
\caption{Mapping between MLE-Bench task names used in our evaluation and their corresponding official task IDs.}
\label{tab:mle-bench-task-mapping}
\begin{tabular}{ll}
    \toprule
    \textbf{Task Name} & \textbf{Task ID} \\ 
    \midrule
    3D Object Detection & \texttt{3d-object-detection-for-autonomous-vehicles} \\
    AI4Code & \texttt{AI4Code} \\
    iMet 2020 FGVC7 & \texttt{imet-2020-fgvc7} \\
    RSNA Brain Tumor & \texttt{rsna-miccai-brain-tumor-radiogenomic-classification} \\ 
    iNaturalist 2019 FGVC6 & \texttt{inaturalist-2019-fgvc6} \\
    \bottomrule
\end{tabular}
\end{table}

We evaluate \sys{} on MLE-Bench and Parameter Golf. Details regarding the benchmarks and our evaluation setup are provided below.

\paragraph{MLE-Bench.} We selected five MLE-Bench~\cite{chan2024mle} Kaggle competitions spanning medical imaging, fine-grained recognition, and 3D perception. Table~\ref{tab:mle-bench-task-mapping} summarizes the mapping between the task names used in our evaluation and their official task IDs. We specifically targeted tasks in the Medium and High difficulty tiers, as they possess sufficient complexity and substantive research value to serve as meaningful benchmarks for automated scientific discovery and paper writing. Specifically, AI4Code, iMet 2020 FGVC7, and iNaturalist 2019 FGVC6 are classified as Medium-difficulty tasks, whereas 3D Object Detection and RSNA Brain Tumor are High-difficulty tasks. The agent is provided with task descriptions and a code execution environment to develop its solutions. The execution environment consists of a compute node provisioned with 8xH100 GPUs, 192 CPU cores, and 1TB of RAM. To assist with formatting, the agent can query a validation tool an unlimited number of times to ensure its submission CSV files are correctly structured. Furthermore, we permit the agent to query the grading server up to 16 times to obtain evaluation scores on the test data. This deviates from the official MLE-Bench protocol, which restricts evaluation to a single submission of the final generated solution. We adopt this modified setup to better simulate real-world Kaggle competitions, where participants iteratively submit solutions to a public leaderboard for feedback during the development phase. While the agent is instructed to iterate on its solutions primarily based on metrics derived from the validation dataset, the grading server is used sparingly to select the best-performing models on the test set. Additionally, this setup is necessary to allow the agent to report accurate final test metrics when generating the paper.

\paragraph{Parameter Golf.} To assess adaptability in a novel, live environment, we evaluate \sys{} on the Parameter Golf competition~\citep{openai2026parametergolf}. This live competition serves as a well-suited AI research task focused on LLM training. It requires participants to train the highest-performing language model under strict constraints: the final artifact must fit within a 16MB size limit and the training process must complete in under 10 minutes on an 8xH100 system. Performance is evaluated by the compression rate---measured in tokenizer-agnostic bits per byte (BPB)--- on the FineWeb validation set. We provide the agent with a sandbox tool to execute solution code to obtain evaluation metrics. Additionally, the agent is provided with a knowledge base containing official leaderboard solutions up to a cutoff date of April 27, 2026. As of this date, the state-of-the-art (SOTA) score was 1.0611, achieved by a solution titled ``BOS-Fixed SmearGate + LQER + SparseAttnGate + 9-Hparam Stack''. The agent is tasked with building upon these existing leaderboard baselines to discover novel techniques capable of further improving the SOTA score.

%% file: sections/012e_baseline_adaptation.tex
%% ============================================================
%% Appendix: Baseline Adaptation Details
%% ============================================================

\section{Baseline Adaptation Details}
\label{app:baseline_adaptation}

All baselines are open-source systems adapted to the ADRS benchmark under identical conditions: Gemini~3.1~Pro backbone, up to 20 solver iterations per task, 2-hour code generation windows, and 3~seeds per task.
Runs that crashed due to infrastructure issues (API timeouts, rate limits, \LaTeX{} compilation errors) were re-attempted with fresh state up to 3~times. No run was re-attempted to improve solver scores.
Below we detail the per-system adaptations.

\paragraph{Sakana AI-Scientist v2 (Sakana).}
Sakana~v2 required the deepest adaptation.
The original codebase assumes an ML-training workflow throughout its 4-stage best-first tree search (BFTS) pipeline: stage goals instruct the agent to ``tune learning rates and batch sizes,'' ``introduce datasets from HuggingFace,'' and ``avoid changing the model architecture.''
%In an initial 15-run trial with unmodified prompts, most runs trained irrelevant PyTorch neural networks instead of writing ADRS optimization code; only \textsc{txn\_scheduling} consistently used simulated annealing with the real evaluator.

To adapt for ADRS, we rewrote all four stage goals in \texttt{agent\_manager.py} and 14~prompt locations in \texttt{parallel\_agent.py}, replacing ML-specific terminology (``training loop,'' ``model architecture,'' ``dataset'') with optimization-domain equivalents (``solution search,'' ``algorithm design,'' ``problem instance'').
%The rewritten stage goals direct Stage~1 toward a working implementation that produces a valid score, Stage~2 toward hyperparameter tuning across problem instances, Stage~3 toward creative algorithmic exploration (hybrid solvers, learned construction policies, domain-specific heuristics), and Stage~4 toward systematic ablation studies.

Beyond prompt changes, we created a bridge entry point (\texttt{perform\_experiments\_adrs.py}) that loads ADRS task specifications, injects the evolve-block starter code, and routes evaluation through the one-ai-scientist task loader rather than Sakana's built-in interpreter.
We also added a Gemini~3.1~Pro backend (\texttt{backend\_gemini.py}), since the original codebase only supports OpenAI-compatible APIs.
%The writeup pipeline required two fixes: (1)~the citation injection regex targeted the ICBINB template's \verb|\begin{filecontents}| block, absent in the NeurIPS template---we switched to direct append to \texttt{references.bib}; and (2)~the NeurIPS template lacked a \verb|\graphicspath| directive, causing figures to render as placeholders.
In total, 16~source files were modified or created, plus 5~task-specific idea files and the NeurIPS~2026 \LaTeX{} template.
%The BFTS pipeline runs with iteration budgets of 20/12/12/18 across stages~1--4, 4~parallel workers, 3~search drafts per stage, max debug depth~3, and a 1-hour execution timeout per iteration.
%The writeup phase gathers citations over 10~rounds from Semantic Scholar, aggregates experiment plots via a VLM, and generates an 8-page NeurIPS-format paper.
% Each Slurm job runs on a CPU-only c3~node (8~vCPUs, 32\,GB~RAM) with a 12-hour wall-time limit.
% Of the 15~runs, 2~had infrastructure failures (one OOM kill, one Slurm time-limit exceeded), both recovered from checkpoints.

Four post-integration fixes were required, discovered during forensic audit:
\begin{enumerate}[nosep,leftmargin=*]
\item \emph{Canonical evaluator injection.}
BFTS bypasses its own \texttt{exec\_callback} and executes generated code directly. The evaluation callback we registered was dead code.
We instead appended a guarded canonical evaluation harness to the initial code template, so every generated \texttt{runfile.py} runs the official ADRS evaluator and prints canonical metrics to stdout for the BFTS metric parser.
\item \emph{Deterministic best-node selection.}
BFTS selects its ``best'' node via an LLM call at temperature$=$0.3 that considers ``training dynamics'' and ``plot quality''---not raw score alone.
We added a deterministic argmax over all nodes across all 4~stages, producing \texttt{best\_solver.py} with a fresh canonical re-evaluation.
\item \emph{Paper--code alignment.}
The writeup pipeline reads the LLM-selected node, while \texttt{best\_solver.py} comes from the deterministic argmax---yielding different algorithms in all 15~runs.
The papers evaluated in this study were not regenerated after this fix, so paper descriptions may refer to the LLM-selected node rather than the highest-scoring one.
\item \emph{Plot generation path.}
The upstream \texttt{load\_exp\_summaries()} hardcodes \texttt{logs/0-run/} as the summary directory, but our bridge writes summaries to the run root.
The mismatch caused \texttt{aggregate\_plots} to receive empty data, producing zero figures across all 15~runs.
All 15~Sakana papers are therefore figure-free.
\end{enumerate}

Sakana's specification violation patterns (10/15 papers flagged) are analyzed in \S\ref{app:coe_audit_details}. The dominant cause is a design mismatch between BFTS's multi-stage architecture (which assumes self-contained evaluation) and ADRS's solver contract (which treats the evaluator as off-limits infrastructure).

\paragraph{AutoResearchClaw (ARC).}
ARC~v0.4.0 (commit \texttt{39c996b}, April~2026) required 2~source patches: one to add Gemini~3.x to the thinking-model whitelist (2~lines in \texttt{llm/client.py}), and one to configure the scaffold directory for ADRS tasks.
No changes were made to ARC's search logic, stopping criteria, or evaluation pipeline.
We set \texttt{max\_iterations=20} (6.7$\times$ ARC's default of~3) and \texttt{max\_refine\_duration=7200s}.
ARC's internal limits (max 2~decision pivots, max 3~repair cycles) are architectural and were not modified.
ARC's post-pipeline scoring (PQEP) failed on 4 of 15~runs due to a regex bug in evaluator path substitution. These were resolved by re-running the canonical ADRS evaluator on the submitted code.
Figure generation failed across all 15~runs because ARC's \texttt{FigureAgent} requires a Docker image not available in our environment. All papers were produced without figures.

ARC's blueprint planner designs multi-file project architectures (\texttt{config.py}, \texttt{models.py}, etc.) before seeing ADRS's single-file constraint.
The scaffold partially compensates by renaming the entry point to \texttt{program.py}, but the solver still imports helper modules that exist only in the original experiment directory.
Re-evaluation in isolation fails for 5 of 15~solvers---the dominant root cause of ARC's I1 and I4 failures.

\paragraph{DeepScientist (DS).}
DS~v1.5.17 (tag \texttt{v1.5.17}, April~2026) on Codex~CLI~0.57.0 required prompt-only changes---no source files were patched.
Task specifications and the NeurIPS~2026 template were injected via the write-phase prompt.
We set \texttt{CODE\_TIMEOUT=7200s} and \texttt{WRITE\_TIMEOUT=5400s}, with up to 3~retries each for code and write phases.
DS's write skill instructs the agent to retrieve citations via Semantic Scholar, arXiv, and CrossRef APIs. However, across all 15~write-phase logs, the agent never called any retrieval API or MCP tool, generating all references from model memory.
This is a model compliance failure---the tools are available but the agent consistently shortcuts the citation retrieval instructions.

\paragraph{AI-Researcher (AIR).}
AIR required the most extensive adaptation: 19~source files were patched to interface with ADRS task configurations, the Docker-based sandbox, and the NeurIPS~2026 paper template.
AIR's MetaChain ReAct agent runs inside a Docker container. We configured the same Gemini~3.1~Pro model and iteration budgets as other systems.
Of the 15~runs, 6~required re-runs due to Gemini API rate limits or sandbox crashes. Four papers required manual \LaTeX{} fixes (trailing newlines, malformed \texttt{filecontents} environments, bibliography corruption).

%% file: main.bbl
\begin{thebibliography}{33}
\providecommand{\natexlab}[1]{#1}
\providecommand{\url}[1]{\texttt{#1}}
\expandafter\ifx\csname urlstyle\endcsname\relax
  \providecommand{\doi}[1]{doi: #1}\else
  \providecommand{\doi}{doi: \begingroup \urlstyle{rm}\Url}\fi

\bibitem[Cemri et~al.(2026)Cemri, Agrawal, Gupta, Liu, Cheng, Mang, Naren, Erdogan, Sen, Zaharia, et~al.]{cemri2026adaevolve}
M.~Cemri, S.~Agrawal, A.~Gupta, S.~Liu, A.~Cheng, Q.~Mang, A.~Naren, L.~E. Erdogan, K.~Sen, M.~Zaharia, et~al.
\newblock {AdaEvolve}: Adaptive {LLM} driven zeroth-order optimization.
\newblock \emph{arXiv preprint arXiv:2602.20133}, 2026.

\bibitem[Chan et~al.(2024)Chan, Chowdhury, Jaffe, Aung, Sherburn, Mays, Starace, Liu, Maksin, Patwardhan, et~al.]{chan2024mle}
J.~S. Chan, N.~Chowdhury, O.~Jaffe, J.~Aung, D.~Sherburn, E.~Mays, G.~Starace, K.~Liu, L.~Maksin, T.~Patwardhan, et~al.
\newblock {MLE-Bench}: Evaluating machine learning agents on machine learning engineering.
\newblock \emph{arXiv preprint arXiv:2410.07095}, 2024.

\bibitem[Chen et~al.(2025)Chen, Anumasa, Lin, Shah, Goyal, and Liu]{zhong2025autobench}
T.~Chen, S.~Anumasa, B.~Lin, V.~Shah, A.~Goyal, and D.~Liu.
\newblock {Auto-Bench}: An automated benchmark for scientific discovery in {LLM}s.
\newblock \emph{arXiv preprint arXiv:2502.15224}, 2025.

\bibitem[Cheng et~al.(2025{\natexlab{a}})Cheng, Liu, Pan, Li, Agarwal, Cemri, Wang, Krentsel, Xia, Park, et~al.]{cheng2025letthebarbarians}
A.~Cheng, S.~Liu, M.~Pan, Z.~Li, S.~Agarwal, M.~Cemri, B.~Wang, A.~Krentsel, T.~Xia, J.~Park, et~al.
\newblock Let the barbarians in: How {AI} can accelerate systems performance research.
\newblock \emph{arXiv preprint arXiv:2512.14806}, 2025{\natexlab{a}}.

\bibitem[Cheng et~al.(2025{\natexlab{b}})Cheng, Liu, Pan, Li, Wang, Krentsel, Xia, Cemri, Park, Yang, et~al.]{cheng2025barbarians}
A.~Cheng, S.~Liu, M.~Pan, Z.~Li, B.~Wang, A.~Krentsel, T.~Xia, M.~Cemri, J.~Park, S.~Yang, et~al.
\newblock Barbarians at the gate: How {AI} is upending systems research.
\newblock \emph{arXiv preprint arXiv:2510.06189}, 2025{\natexlab{b}}.

\bibitem[Goyal et~al.(2026)Goyal, Parmar, Song, Palangi, Pfister, and Yoon]{goyal2026scholarpeer}
P.~Goyal, M.~Parmar, Y.~Song, H.~Palangi, T.~Pfister, and J.~Yoon.
\newblock {ScholarPeer}: A context-aware multi-agent framework for automated peer review.
\newblock \emph{arXiv preprint arXiv:2601.22638}, 2026.

\bibitem[H{\"a}rder and Reuter(1983)]{haerder1983principles}
T.~H{\"a}rder and A.~Reuter.
\newblock Principles of transaction-oriented database recovery.
\newblock \emph{{ACM} Computing Surveys}, 15\penalty0 (4):\penalty0 287--317, 1983.

\bibitem[Huang et~al.(2023)Huang, Vora, Liang, and Leskovec]{huang2024mlagentbench}
Q.~Huang, J.~Vora, P.~Liang, and J.~Leskovec.
\newblock {MLAgentBench}: Evaluating language agents on machine learning experimentation.
\newblock \emph{arXiv preprint arXiv:2310.03302}, 2023.

\bibitem[Jansen et~al.(2025)Jansen, Tafjord, Radensky, Siangliulue, Hope, Dalvi, Majumder, Weld, and Clark]{jansen2025codescientist}
P.~Jansen, O.~Tafjord, M.~Radensky, P.~Siangliulue, T.~Hope, B.~Dalvi, B.~P. Majumder, D.~S. Weld, and P.~Clark.
\newblock {CodeScientist}: End-to-end semi-automated scientific discovery with code-based experimentation.
\newblock In \emph{Findings of the Association for Computational Linguistics: ACL 2025}, pages 13370--13467, 2025.

\bibitem[Kon et~al.(2025{\natexlab{a}})Kon, Liu, Ding, Qiu, Yang, Huang, Srinivasa, Lee, Chowdhury, and Chen]{kon2025curie}
P.~T.~J. Kon, J.~Liu, Q.~Ding, Y.~Qiu, Z.~Yang, Y.~Huang, J.~Srinivasa, M.~Lee, M.~Chowdhury, and A.~Chen.
\newblock {Curie}: Toward rigorous and automated scientific experimentation with {AI} agents.
\newblock \emph{arXiv preprint arXiv:2502.16069}, 2025{\natexlab{a}}.

\bibitem[Kon et~al.(2025{\natexlab{b}})Kon, Liu, Zhu, Ding, Peng, Xing, Huang, Qiu, Srinivasa, Lee, et~al.]{weng2025expbench}
P.~T.~J. Kon, J.~Liu, X.~Zhu, Q.~Ding, J.~Peng, J.~Xing, Y.~Huang, Y.~Qiu, J.~Srinivasa, M.~Lee, et~al.
\newblock {EXP-Bench}: Can {AI} conduct {AI} research experiments?
\newblock \emph{arXiv preprint arXiv:2505.24785}, 2025{\natexlab{b}}.

\bibitem[Liu et~al.(2026{\natexlab{a}})Liu, Qiu, Li, Li, Ji, Han, Ye, Xia, Dong, Zhang, et~al.]{li2024autoresearchclaw}
J.~Liu, S.~Qiu, M.~Li, B.~Li, H.~Ji, S.~Han, X.~Ye, P.~Xia, Z.~Dong, C.~Zhang, et~al.
\newblock Autoresearchclaw: Self-reinforcing autonomous research with human-ai collaboration.
\newblock \emph{arXiv preprint arXiv:2605.20025}, 2026{\natexlab{a}}.

\bibitem[Liu et~al.(2023{\natexlab{a}})Liu, Zhang, and Liang]{liu2023evaluating}
N.~F. Liu, T.~Zhang, and P.~Liang.
\newblock Evaluating verifiability in generative search engines.
\newblock In \emph{Findings of the Association for Computational Linguistics: EMNLP 2023}, pages 7001--7025, 2023{\natexlab{a}}.

\bibitem[Liu et~al.(2024)Liu, Lin, Hewitt, Paranjape, Bevilacqua, Petroni, and Liang]{liu2024lost}
N.~F. Liu, K.~Lin, J.~Hewitt, A.~Paranjape, M.~Bevilacqua, F.~Petroni, and P.~Liang.
\newblock Lost in the middle: How language models use long contexts.
\newblock \emph{Transactions of the Association for Computational Linguistics}, 12:\penalty0 157--173, 2024.

\bibitem[Liu et~al.(2026{\natexlab{b}})Liu, Agarwal, Maheswaran, Cemri, Li, Mang, Naren, Boneh, Cheng, Pan, et~al.]{liu2026evox}
S.~Liu, S.~Agarwal, M.~Maheswaran, M.~Cemri, Z.~Li, Q.~Mang, A.~Naren, E.~Boneh, A.~Cheng, M.~Z. Pan, et~al.
\newblock {EvoX}: Meta-evolution for automated discovery.
\newblock \emph{arXiv preprint arXiv:2602.23413}, 2026{\natexlab{b}}.

\bibitem[Liu et~al.(2026{\natexlab{c}})Liu, Cemri, Agarwal, Krentsel, Naren, Mang, Li, Gupta, Maheswaran, Cheng, Pan, Boneh, Ramchandran, Sen, Dimakis, Zaharia, and Stoica]{skydiscover2026}
S.~Liu, M.~Cemri, S.~Agarwal, A.~Krentsel, A.~Naren, Q.~Mang, Z.~Li, A.~Gupta, M.~Maheswaran, A.~Cheng, M.~Pan, E.~Boneh, K.~Ramchandran, K.~Sen, A.~G. Dimakis, M.~Zaharia, and I.~Stoica.
\newblock Skydiscover: A flexible framework for ai-driven scientific and algorithmic discovery, 2026{\natexlab{c}}.
\newblock URL \url{https://skydiscover-ai.github.io/blog.html}.

\bibitem[Liu et~al.(2023{\natexlab{b}})Liu, Yu, Zhang, Xu, Lei, Lai, Gu, Ding, Men, Yang, et~al.]{liu2023agentbench}
X.~Liu, H.~Yu, H.~Zhang, Y.~Xu, X.~Lei, H.~Lai, Y.~Gu, H.~Ding, K.~Men, K.~Yang, et~al.
\newblock {AgentBench}: Evaluating {LLM}s as agents.
\newblock \emph{arXiv preprint arXiv:2308.03688}, 2023{\natexlab{b}}.

\bibitem[Liu et~al.(2025)Liu, Yang, Xie, Ni, Gao, Li, Tang, Ouyang, Cambria, and Zhou]{liu2025researchbench}
Y.~Liu, Z.~Yang, T.~Xie, J.~Ni, B.~Gao, Y.~Li, S.~Tang, W.~Ouyang, E.~Cambria, and D.~Zhou.
\newblock {ResearchBench}: Benchmarking {LLM}s in scientific discovery via inspiration-based task decomposition.
\newblock \emph{arXiv preprint arXiv:2503.21248}, 2025.

\bibitem[Lu et~al.(2024)Lu, Lu, Lange, Foerster, Clune, and Ha]{lu2024aiscientist}
C.~Lu, C.~Lu, R.~T. Lange, J.~Foerster, J.~Clune, and D.~Ha.
\newblock The {AI} scientist: Towards fully automated open-ended scientific discovery.
\newblock \emph{arXiv preprint arXiv:2408.06292}, 2024.

\bibitem[Lupidi et~al.(2026)Lupidi, Gauri, Foster, Omari, Magka, Pepe, Audran-Reiss, Aghamelu, Baldwin, Cipolina-Kun, et~al.]{jansen2025airsbench}
A.~Lupidi, B.~Gauri, T.~S. Foster, B.~A. Omari, D.~Magka, A.~Pepe, A.~Audran-Reiss, M.~Aghamelu, N.~Baldwin, L.~Cipolina-Kun, et~al.
\newblock {AIRS-Bench}: A suite of tasks for frontier {AI} research science agents.
\newblock \emph{arXiv preprint arXiv:2602.06855}, 2026.

\bibitem[Lyu et~al.(2026)Lyu, Zhang, Yi, Zhao, Guo, Hu, Piotrowski, Kaliski, Urbani, Meng, et~al.]{li2026evoscientist}
Y.~Lyu, X.~Zhang, X.~Yi, Y.~Zhao, S.~Guo, W.~Hu, J.~Piotrowski, J.~Kaliski, J.~Urbani, Z.~Meng, et~al.
\newblock {EvoScientist}: Towards multi-agent evolving {AI} scientists for end-to-end scientific discovery.
\newblock \emph{arXiv preprint arXiv:2603.08127}, 2026.

\bibitem[Min et~al.(2023)Min, Krishna, Lyu, Lewis, Yih, Koh, Iyyer, Zettlemoyer, and Hajishirzi]{min2023factscore}
S.~Min, K.~Krishna, X.~Lyu, M.~Lewis, W.-t. Yih, P.~Koh, M.~Iyyer, L.~Zettlemoyer, and H.~Hajishirzi.
\newblock {FActScore}: Fine-grained atomic evaluation of factual precision in long form text generation.
\newblock In \emph{Proceedings of the 2023 Conference on Empirical Methods in Natural Language Processing}, pages 12076--12100, 2023.

\bibitem[Novikov et~al.(2025)Novikov, V{\~u}, Eisenberger, Dupont, Huang, Wagner, Shirobokov, Kozlovskii, Ruiz, Mehrabian, et~al.]{novikov2025alphaevolve}
A.~Novikov, N.~V{\~u}, M.~Eisenberger, E.~Dupont, P.-S. Huang, A.~Z. Wagner, S.~Shirobokov, B.~Kozlovskii, F.~J. Ruiz, A.~Mehrabian, et~al.
\newblock {AlphaEvolve}: A coding agent for scientific and algorithmic discovery.
\newblock \emph{arXiv preprint arXiv:2506.13131}, 2025.

\bibitem[OpenAI(2026)]{openai2026parametergolf}
OpenAI.
\newblock Parameter golf: {OpenAI} model craft challenge.
\newblock \url{https://github.com/openai/parameter-golf}, 2026.

\bibitem[Press et~al.(2024)Press, Hochlehnert, Prabhu, Udandarao, Press, and Bethge]{press2024citeme}
O.~Press, A.~Hochlehnert, A.~Prabhu, V.~Udandarao, O.~Press, and M.~Bethge.
\newblock {CiteME}: Can language models accurately cite scientific claims?
\newblock \emph{Advances in Neural Information Processing Systems}, 37:\penalty0 7847--7877, 2024.

\bibitem[Pu et~al.(2025)Pu, Lin, and Chen]{wu2025piflow}
Y.~Pu, T.~Lin, and H.~Chen.
\newblock {PiFlow}: Principle-aware scientific discovery with multi-agent collaboration.
\newblock \emph{arXiv preprint arXiv:2505.15047}, 2025.

\bibitem[Schmidgall et~al.(2025)Schmidgall, Su, Wang, Sun, Wu, Yu, Liu, Moor, Liu, and Barsoum]{schmidgall2025agentlab}
S.~Schmidgall, Y.~Su, Z.~Wang, X.~Sun, J.~Wu, X.~Yu, J.~Liu, M.~Moor, Z.~Liu, and E.~Barsoum.
\newblock Agent laboratory: Using {LLM} agents as research assistants.
\newblock \emph{Findings of the Association for Computational Linguistics: EMNLP 2025}, pages 5977--6043, 2025.

\bibitem[Starace et~al.(2025)Starace, Jaffe, Sherburn, Aung, Chan, Maksin, Dias, Mays, Kinsella, Thompson, et~al.]{starace2025paperbench}
G.~Starace, O.~Jaffe, D.~Sherburn, J.~Aung, J.~S. Chan, L.~Maksin, R.~Dias, E.~Mays, B.~Kinsella, W.~Thompson, et~al.
\newblock {PaperBench}: Evaluating {AI}'s ability to replicate {AI} research.
\newblock \emph{arXiv preprint arXiv:2504.01848}, 2025.

\bibitem[Tang et~al.(2025)Tang, Xia, Li, and Huang]{tang2025airesearcher}
J.~Tang, L.~Xia, Z.~Li, and C.~Huang.
\newblock {AI-Researcher}: Autonomous scientific innovation.
\newblock \emph{arXiv preprint arXiv:2505.18705}, 2025.

\bibitem[Wang et~al.(2026)Wang, Bai, Luo, Su, Sun, Yu, Liu, Zhou, Cardie, Dredze, et~al.]{shojaee2025firebench}
Z.~Wang, F.~Bai, Z.~Luo, J.~Su, K.~Sun, X.~Yu, J.~Liu, K.~Zhou, C.~Cardie, M.~Dredze, et~al.
\newblock {FIRE-Bench}: Evaluating agents on the rediscovery of scientific insights.
\newblock \emph{arXiv preprint arXiv:2602.02905}, 2026.

\bibitem[Weng et~al.(2025)Weng, Zhu, Xie, Sun, Lin, Liu, and Zhang]{huang2025deepscientist}
Y.~Weng, M.~Zhu, Q.~Xie, Q.~Sun, Z.~Lin, S.~Liu, and Y.~Zhang.
\newblock Deepscientist: Advancing frontier-pushing scientific findings progressively.
\newblock \emph{arXiv preprint arXiv:2509.26603}, 2025.

\bibitem[Xu et~al.(2025)Xu, Lu, Ye, Hu, and Liu]{xu2025researcherbench}
T.~Xu, P.~Lu, L.~Ye, X.~Hu, and P.~Liu.
\newblock {ResearcherBench}: Evaluating deep {AI} research systems on the frontiers of scientific inquiry.
\newblock \emph{arXiv preprint arXiv:2507.16280}, 2025.

\bibitem[Yamada et~al.(2025)Yamada, Lange, Lu, Hu, Lu, Foerster, Clune, and Ha]{yamada2025aiscientistv2}
Y.~Yamada, R.~T. Lange, C.~Lu, S.~Hu, C.~Lu, J.~Foerster, J.~Clune, and D.~Ha.
\newblock The {AI} scientist-v2: Workshop-level automated scientific discovery via agentic tree search.
\newblock \emph{arXiv preprint arXiv:2504.08066}, 2025.

\end{thebibliography}
